\documentclass[10pt,twocolumn,letterpaper]{article}

 \usepackage{cvpr}              

\usepackage[dvipsnames]{xcolor}

\usepackage{xcolor}

\newcommand{\ti}{{\emph{T2I}~}}
\newcommand{\newpara}[1]{{\vspace{0.5em}\newline\textbf{#1~}}}
\usepackage{amsmath}
\usepackage{amssymb}
\usepackage{mathtools}
\usepackage{amsthm}
\usepackage{microtype}
\usepackage{multicol}
\usepackage{booktabs}
\usepackage{multirow}
\usepackage{pifont}
\usepackage{algpseudocode}
\usepackage{algorithm}
\usepackage{wrapfig}
\usepackage{tabularray}

\long\def\comment#1{}
\definecolor{cvprblue}{rgb}{0.21,0.49,0.74}
\usepackage[pagebackref,breaklinks,colorlinks,citecolor=cvprblue]{hyperref}

\def\paperID{4885} 
\def\confName{CVPR}
\def\confYear{2024}

\title{Text-guided Explorable Image Super-resolution}

\author{Kanchana Vaishnavi Gandikota\textsuperscript{*}
\qquad
Paramanand Chandramouli\textsuperscript{*}\\
\centering{
Institute for Vision and Graphics, University of Siegen}
}
\begin{document}
\maketitle
\begin{figure*}
		\centering
    \small   
    \includegraphics[width=0.33\linewidth]{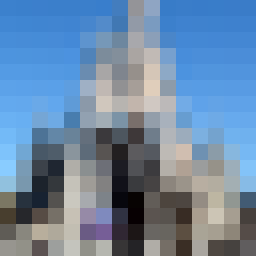}
    \includegraphics[width=0.33\linewidth]{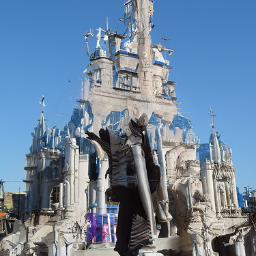}
    \includegraphics[width=0.33\linewidth]{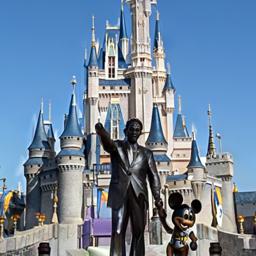}\\
     \resizebox{\linewidth}{!}{\begin{tabular}{ccc}
     \phantom{retsrzdtigujiserdtf}  LR input\phantom{retsrzdtigujis}   & DPS \cite{chung2022diffusion} using diffusion model trained on Imagenet\phantom{retsrzig} &\phantom{retsrzdtj}  \textbf{Ours}\phantom{retsrzdtigujiserdt} 
\end{tabular}}
    'A statue of Walt Disney holding Mickey Mouse hands is showing in front of Cinderellas castle.'\\
   \resizebox{\linewidth}{!}
{\begin{tabular}{cc}
  \multirow{2}{*}{
  \begin{tabular}{c}  
  \includegraphics[width=0.15\linewidth]{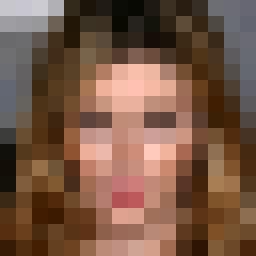}\\
  LR input\\
 \end{tabular}}&\hspace{-11pt}
      \includegraphics[width=0.21\linewidth]{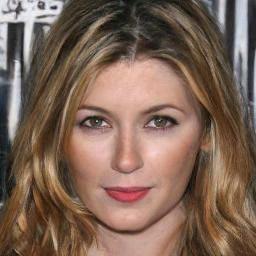} 
      \includegraphics[width=0.21\linewidth]{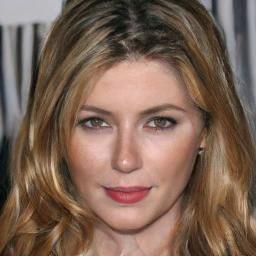}       
      \includegraphics[width=0.21\linewidth]{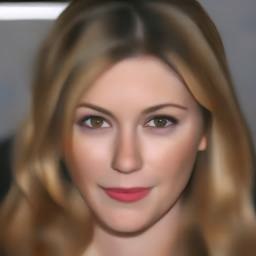}
      \includegraphics[width=0.21\linewidth]{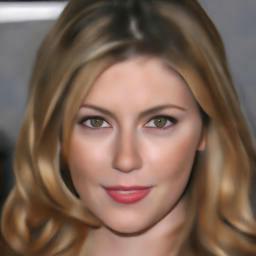}\\
      &
$\xleftarrow{\hspace*{0.15\linewidth}}\text{DPS \cite{chung2022diffusion}} \xrightarrow{\hspace*{0.15\linewidth}}\xleftarrow{\hspace*{0.15\linewidth}}\text{DDNM \cite{wang2022zero}} \xrightarrow{\hspace*{0.15\linewidth}}$\\
     & \hspace{-9pt}      \includegraphics[width=0.21\linewidth]{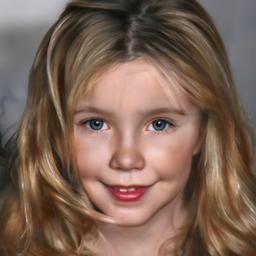}       \includegraphics[width=0.21\linewidth]{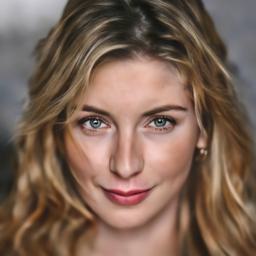} 
      \includegraphics[width=0.21\linewidth]{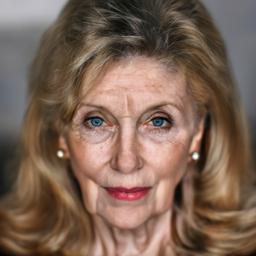}
      \includegraphics[width=0.21\linewidth]{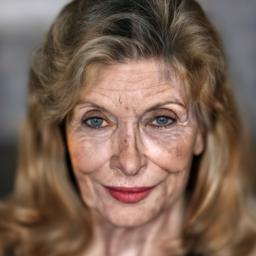}\\
      &'A smiling girl'\phantom{retsrzdtiigujj} 'a smiling woman'\phantom{retsrzdtiguj} 'an elderly woman'\phantom{retsrzdti} 'smiling elderly woman'
      \end{tabular}}    
    		\caption{
		\textbf{Text guided image Super-resolution.} 
		 We explore consistent reconstructions to image super-resolution problems through text prompts 
   while achieving perfect data consistency with the given inputs for all solutions. Shown are a) extreme super-resolution of natural images (top), b) face super-resolution (bottom), with an upsampling factor of 16.}
		\label{fig:teaser}	

\end{figure*}
\begin{abstract}

 In this paper, we introduce the problem of zero-shot text-guided exploration of the solutions to open-domain image super-resolution. Our goal is to allow users to explore diverse, semantically accurate reconstructions that preserve data consistency with the low-resolution inputs for different large downsampling factors without explicitly training for these specific degradations. 
 We propose two approaches for zero-shot text-guided super-resolution -  i) modifying the generative process of text-to-image (\ti) diffusion models to promote consistency with low-resolution inputs, and ii) incorporating language guidance into zero-shot diffusion-based restoration methods. We show that the proposed approaches result in diverse solutions that match the semantic meaning provided by the text prompt while preserving data consistency with the degraded inputs. We evaluate the proposed baselines for the task of extreme super-resolution and demonstrate advantages in terms of restoration quality, diversity, and explorability of solutions. 
\end{abstract} 
\let\thefootnote\relax\footnotetext{* indicates the authors contributed equally.}
\section{Introduction}
 The goal of image super-resolution is to recover a high-quality image, given a low-resolution (LR) observation $\mathbf{y}$,
\begin{equation}
\mathbf{y} = \mathbf{A}\mathbf{x} + \mathbf{n},
\label{eq:measurement}
\end{equation}
where  $\mathbf{A}$, $\mathbf{x}$, and $\mathbf{n}$ represent the down-sampling operator, ground truth image, and measurement noise respectively. Image super-resolution is highly ill-posed, especially at large super-resolution factors with many valid solutions satisfying the data consistency accurately. While most of the recent state-of-the-art supervised deep networks for super-resolution \cite{jo2020investigating,chan2021glean,wang2022panini} recover only a single image from this solution space,  there are also methods utilizing conditional or unconditional generative models \cite{bahat2020explorable,buhler2020deepsee,lugmayr2020srflow,li2022srdiff,kawar2022denoising,wang2022zero}, which allow sampling multiple solutions. A few of these works also allow exploring the solution space, using graphical user inputs  \cite{bahat2020explorable} or semantic maps \cite{buhler2020deepsee}. Yet, even these methods are tailored to images of specific classes such as faces, or trained for specific super-resolution factors. On the other hand, natural language provides a simpler and more intuitive means of conveying semantic concepts. For instance, it is easier to provide detailed descriptions or convey concepts such as age, gender, emotion, and race through text rather than graphical inputs alone.  Therefore, a method that guides image super-resolution through text can greatly aid the exploration of semantically meaningful solutions.

In this paper, we propose for the first time zero-shot open-domain image super-resolution using simple and intuitive text prompts. Our goal is to explore via text prompts, diverse and semantically accurate reconstructions that preserve data consistency with the low-resolution inputs for different large downsampling factors without explicitly training for these specific degradations. Towards this goal, we exploit recent advances in text-to-image (\ti) generative models \cite{saharia2022photorealistic,ramesh2022hierarchical,rombach2022high}, contrastive language image pretraining (CLIP)~\cite{radford2021learning}, and unsupervised zero-shot approaches to image recovery using diffusion-based generative models \cite{chung2022diffusion,song2023pseudoinverseguided,wang2022zero}. We explore two paradigms for zero-shot text-guided super-resolution. In the first approach, we adapt recent diffusion-based zero-shot super-resolution approaches to \ti models by appropriately modifying the generative process. We consider recent state-of-the-art text-to-image diffusion models, open-sourced versions of DALL-e2 \cite{ramesh2022hierarchical}, and Imagen \cite{saharia2022photorealistic},   and adapt these models for zero-shot super-resolution using different zero-shot approaches for diffusion-based image recovery \cite{song2023pseudoinverseguided,chung2022diffusion,wang2022zero}. As these \ti models comprise of a cascade of diffusion models at different resolutions, we modify the zero-shot approaches accordingly to deal with multi-stage generation. In the second approach, we modify an existing diffusion-based zero-shot restoration approach to incorporate additional language guidance through  CLIP. We focus on extreme image super-resolution with large upscale factors, as this problem is severely ill-posed, and allows exploration of a larger solution space.

Fig.~\ref{fig:teaser} illustrates the benefits of text guidance in extreme super-resolution. Existing zero-shot methods such as \cite{chung2022diffusion, wang2020understanding} cannot recover realistic details when the ground truth has complex content as seen in the super-resolution results of \cite{chung2022diffusion} on challenging input. On the other hand, the use of text enables the recovery of complex scene content with specific details matching the text prompt. The use of text also effortlessly improves diversity in solutions for face super-resolution in terms of age, expression, gender, race, and other attributes over \cite{chung2022diffusion,wang2022zero} which recover images with limited diversity. 
We evaluate the proposed baselines in terms of realism, fidelity with low-resolution input, and agreement with text, in several qualitative, quantitative, and human evaluations. Extensive experimental evaluations demonstrate the benefit of using zero-shot text guidance in terms of flexibility,  diversity, and explorability of solutions to extreme super-resolution. 
  Our work opens up a promising direction of developing efficient tools for text-guided exploration of image recovery.
\section{Preliminaries}
\subsection{Denoising Diffusion Probabilistic Models (DDPM)} DDPM generative models  \cite{ho2020denoising} employ two diffusion processes: \emph{i)~A  forward process} slowly noising a data sample $\boldsymbol{x}_0$ into Gaussian distribution $\mathcal{N}$  in $T$ steps, with the evolution of a sample $\mathbf{x}_{t}$ at time-step $t$ given by:
\begin{align}
\begin{split}
    q(\mathbf{x}_{t}|\mathbf{x}_{t-1}):=\mathcal{N}(\mathbf{x}_{t};\sqrt{1-\beta_{t}}\mathbf{x}_{t-1},\beta_{t}\mathbf{I})\\ i.e.,\quad \mathbf{x}_{t}=\sqrt{1-\beta_{t}}\mathbf{x}_{t-1} + \sqrt{\beta_{t}}\boldsymbol{\epsilon}, \quad\boldsymbol{\epsilon}\sim \mathcal{N}(0,\mathbf{I}), 
    \label{eq:ddpm forward 1}
    \end{split}
\end{align}
where ${\{\beta_t\}^T_{t=0}}$  is the noise variance schedule. 
\emph{ii) A learned reverse process} using iterative denoising to generate  samples from the training data distribution $q(\mathbf{x})$ in $T$ steps given by:
\begin{equation}
\begin{gathered}
    p(\mathbf{x}_{t-1}|\mathbf{x}_{t},\mathbf{x}_{0}):=
    \mathcal{N}(\mathbf{x}_{t-1};\boldsymbol{\mu}_{t}(\mathbf{x}_{t},\mathbf{x}_{0}),\sigma_{t}^2\mathbf{I}),\enskip\text{ where,}\\
    \boldsymbol{\mu}_{t}(\mathbf{x}_{t},\mathbf{x}_{0})=\frac{1}{\sqrt{\alpha_{t}}}	\left( \mathbf{x}_{t} - \boldsymbol{\epsilon_\theta}(\mathbf{x}_{t},t)\frac{1-\alpha_{t}}{\sqrt{1-\Bar{\alpha}_{t}}} \right),\enskip\text{ and,}\\
    \mathbf{\sigma}_{t}^{2}=\frac{1-\Bar{\alpha}_{t-1}}{1-\Bar{\alpha}_{t}}\beta_{t},\comment{.\\i.e.,\enskip
    \mathbf{x}_{t-1} =\boldsymbol{\mu}_{t}(\mathbf{x}_{t},\mathbf{x}_{0})+\mathbf{\sigma}_{t}\boldsymbol{z}\enskip\text{ with}\enskip\boldsymbol{z}\sim \mathcal{N}(0,\mathbf{I}).}
    \text{\enskip with}\enskip\alpha_{t} = 1- \beta_{t},~\text{and}~\Bar{\alpha}_{t} = \prod_{i=0}^{t}\alpha_{i},
    \label{eq:ddpm reverse 1}
\end{gathered}
\end{equation}
and  $\boldsymbol{\epsilon_\theta}$ is the learned neural network noise approximator. 
\subsection{Range-Null Space Decomposition}
\noindent When there is no measurement noise in \eqref{eq:measurement}, i.e. $\mathbf{y} = \mathbf{A}\mathbf{x}$,  pseudoinverse operation  $\mathbf{A^{\dagger}}\mathbf{y}$ produces the minimum norm solution with perfect data consistency. Any other solution $(\mathbf{A^{\dagger}}\mathbf{y}+\mathbf{x}_\delta)$ is also data consistent, as long as $\mathbf{x}_\delta$ lies in the null space of $\mathbf{A}$. Note that $\mathbf{x}$ can be decomposed as:
\begin{equation}
    \mathbf{x}\equiv\mathbf{A^{\dagger}}\mathbf{A}\mathbf{x} + (\mathbf{I} - \mathbf{A^{\dagger}}\mathbf{A})\mathbf{x}.
    \label{eq:rnd}
\end{equation}
The component $(\mathbf{I} - \mathbf{A^{\dagger}}\mathbf{A})\mathbf{x}$ is in the null space of $\mathbf{A}$, (with $\mathbf{A}(\mathbf{I} - \mathbf{A^{\dagger}}\mathbf{A})\mathbf{x}\equiv\boldsymbol{0}$). Given an approximate solution $\bar{\mathbf{x}}$, Eq.~\ref{eq:rnd} can be used to construct a data consistent solution \cite{Schwab2018DeepNS,bahat2020explorable,wang2022panini,wang2022zero} given by $\hat{\textbf{x}}$ as,
\begin{equation}
\hat{\mathbf{x}}=\mathbf{A^{\dagger}}\mathbf{y} + (\mathbf{I} - \mathbf{A^{\dagger}}\mathbf{A})\bar{\mathbf{x}}.
    \label{eq:cem_rnd}
\end{equation}
\subsection{Zero-Shot Restoration using Diffusion Models}\label{sec:prelim_restore} We now describe diffusion-based zero-shot restoration methods that we explore in this paper for text-guided restoration. Given a noisy sample $\mathbf{x}_{t}$ at step $t$, \cite{song2020denoising} obtain the clean estimate of $\mathbf{x}_{0}$ as:
\begin{align}
      \mathbf{x}_{0|t} := \frac{1}{\sqrt{\Bar{\alpha}_{t}}}	\left( \mathbf{x}_{t} - \boldsymbol{\epsilon_\theta}(\mathbf{x}_{t},t)\sqrt{1-\Bar{\alpha}_{t}} \right).
    \label{eq:x0t}
\end{align}
The works \citep{chung2022diffusion, song2023pseudoinverseguided}   also utilize $\mathbf{x}_{0|t}$ of a similar form   by  approximating $p(\mathbf{x}_{0}|\mathbf{x}_{t})$ as a  simple Gaussian distribution.
This estimate is used in a guidance function or in a projection to incorporate measurement consistency in the following methods. \vspace{0.5em}\\
\textbf{Diffusion Posterior Sampling} DPS \cite{chung2022diffusion} utilize $\mathbf{x}_{0|t}$ in  reconstruction guidance as:
\begin{align}
\mathbf{x}_{t-1} \gets \mathbf{x}'_{t-1} -  {\rho_t}\nabla_{\mathbf{x}_t}\|\mathbf{y} - \mathbf{A}(\mathbf{x}_{0|t})\|_2^2,
\label{eq:dps_guidance}
\end{align}
where, the intermediate estimate at the previous step $\mathbf{x}'_{t-1}$ is obtained using the usual reverse step, and the gradient of the reconstruction loss for $\|\mathbf{y} - \mathbf{A}(\mathbf{x}_{0|t})\|_2^2$ with respect to ${\mathbf{x}_t}$ is obtained by differentiating through the diffusion model.\vspace{0.5em}\\
\textbf{Pseudoinverse-Guided Diffusion Models }$\Pi$GDM~\citep{song2023pseudoinverseguided} modify \eqref{eq:dps_guidance}   by inverting the measurement model using pseudoinverse as:
\begin{align}
\mathbf{x}_{t-1} \gets \mathbf{x}'_{t-1} -  {\rho_t}\nabla_{\mathbf{x}_t}\| \mathbf{A^{\dagger}}\mathbf{y} -  \mathbf{A^{\dagger}}\mathbf{A}(\mathbf{x}_{0|t})\|_2^2,
\label{eq:pgd_guidance}
\end{align}
and demonstrate improved reconstructions with a reduced number of diffusion steps.\vspace{0.5em}\\
\textbf{Denoising Diffusion Null-Space Models} DDNM  \cite{wang2022zero} utilize the range space-null space decomposition in the reverse diffusion process to rectify the clean estimate at each step to satisfy data consistency as: 
\begin{equation}
    \hat{\mathbf{x}}_{0|t}:=\mathbf{A^{\dagger}}\mathbf{y} + (\mathbf{I} - \mathbf{A^{\dagger}}\mathbf{A})\mathbf{x}_{0|t}.
    \label{eq:ndm core}
\end{equation}
This rectified data consistent estimate $\hat{\mathbf{x}}_{0|t}$ is used in subsequent sampling from $p(\mathbf{x}_{t-1}|\mathbf{x}_{t},\hat{\mathbf{x}}_{0|t})$ in \cite{wang2022zero}. Compared to \cite{chung2022diffusion,song2023pseudoinverseguided}, this method is faster as it does not require differentiation through diffusion model weights.
\subsection{Text guided Image Generation with Diffusion Models}
There are two approaches for text-guided image generation using diffusion models- training text-conditioned diffusion models, or incorporating text guidance into unconditional models using vision-language models such as CLIP \cite{radford2021learning}. 
\subsubsection{Text-to-Image \ti Diffusion Models}
We now describe the \ti generative models we employ for text-guided super-resolution.
\newpara{DALL E-2 unCLIP} \cite{ramesh2022hierarchical}  consists of: 
\emph{i)~a diffusion-based prior} to produce CLIP image embeddings \cite{radford2021learning} from encodings of the input prompt, \emph{ii) a conditional diffusion-based decoder}  to generate images conditioned on CLIP image embeddings and text prompts in a down-sampled pixel space, and
\emph{iii) a diffusion-based super-resolution module}  to upsample the decoder output into a high-resolution image.
\newpara{Imagen} \cite{saharia2022photorealistic} utilizes a pretrained text encoder \cite{raffel2020exploring} to generate embeddings from input text which condition a cascade of
conditional diffusion models to generate images of increasing resolutions. Different from unCLIP, Imagen uses only text embeddings which are used to condition every stage of image generation and super-resolution. 

We consider unCLIP and Imagen diffusion models with two stages, $\boldsymbol{\epsilon_\theta}$ operating on a down-sampled pixel space at resolution $64\times64$, and an upsampling stage $\boldsymbol{\zeta_\theta}$ operating at resolution $256\times256$.
\subsubsection{Training-free text-guided generation}
The idea of training free guidance \cite{yu2023freedom,universal2023} is to incorporate desired conditioning signal $c$ into generation process through appropriate energy function $E$ which measures the distance between desired condition and clean estimate $\mathbf{x}_{0|t}$ at every diffusion step:
\begin{align}
\mathbf{x}_{t-1} \gets \mathbf{x}'_{t-1} -  {\rho_t}\nabla_{\mathbf{x}_t}E(c,\mathbf{x}_{0|t}).
\label{eq:train-free-guid}
\end{align}
For text-guided generation, $E$ can be defined using the distance between the text and image embeddings obtained through CLIP text and image encoders.

\section{Methodology}\label{sec:methodology}
Given a low-resolution image ${y}$ with known degradation operator $A$, our goal is to generate data-consistent solutions $u$ whose attributes can be varied using input text prompts $c$.
\begin{equation}
    \begin{split}
    \textit{Data Consistency}: \quad A\hat{u} \equiv {f},\\\textit{Semantic Consistency}: \quad \hat{u} \sim q({u|c}),
    \label{eq:consistency}
\end{split}
\end{equation}
where $q({u|c})$ denotes the distribution of images $u$ with semantic meaning provided by the text prompt ${c}$. 
To obtain solutions satisfying semantic meaning provided by text prompt as well as measurement consistency, we explore the following methods:
\begin{enumerate}
\item  Zero-shot super-resolution using \ti models.  
\item Incorporating CLIP guidance into zero-shot diffusion-based restoration.
\end{enumerate}
\subsection{Text Guided Super-resolution using \bf{\ti}~Models}
 We consider two recent diffusion-based text-to-image (T2I) generative models that operate in the pixel domain  DALL E-2 \cite{ramesh2022hierarchical} and Imagen \cite{saharia2022photorealistic}.  As these models employ a multi-stage generation process, first in down-sampled pixel-space, followed by upsampling stages, we correspondingly modify the sampling process to incorporate guidance or null-space consistency in both stages of the generation.
 Let $\mathbf{c_1}$ and $\mathbf{c_2}$ denote the conditioning signals in the two stages, $\boldsymbol{\epsilon_\theta}$ and $\boldsymbol{\zeta_\theta}$ denote conditioned diffusion models in the down-sampled stage, and the super-resolution stage respectively. The current estimate of the clean image at each step in the first stage and second stage are respectively given by,
\begin{align}
    \mathbf{x}_{{LR}_{0|t}} = \frac{1}{\sqrt{\Bar{\alpha}_{t}}}	\left( \mathbf{x}_{LR_t} - \boldsymbol{\epsilon_\theta}(\mathbf{x}_{LR_t},t|\mathbf{c}_1)\sqrt{1-\Bar{\alpha}_{t}} \right),     
    \label{eq:xlr0t}
\end{align}
\begin{equation}
\mathbf{x}_{0|t} = \frac{1}{\sqrt{\Bar{\alpha}}_{t}}\left( \mathbf{x}_{t} - \boldsymbol{\zeta_\theta}(\mathbf{x}_{t},t|\mathbf{c}_2)\sqrt{1-\Bar{\alpha}_{t}} \right).
\label{eq:xhr0t}
\end{equation}
For Imagen, $\mathbf{c}_1$ corresponds to text embeddings from the text encoder, and $\mathbf{c}_2$ contains $\mathbf{x}_{LR}$, in addition to text embeddings. For unCLIP $\mathbf{c}_1$ corresponds to a combination of  CLIP image embeddings produced by the prior model and text embeddings, and $\mathbf{c}_2$ is the output $\mathbf{x}_{LR}$ of the first stage.

We first recover a lower resolution version $\mathbf{x}_{LR}$ by using a modified measurement $\mathbf{A}_{LR}$ which takes into account the downsampling operation for text-conditioned diffusion in low resolution. For the subsequent super-resolution using the model $\zeta_\theta$, we consider the actual measurement operator $\mathbf{A}$.  We adapt zero-shot methods discussed in Sec.~\ref{sec:prelim_restore} as follows: 
\paragraph{\ti-DPS:}  We incorporate reconstruction guidance in both stages. In down-sampled pixel space, reconstruction guidance is given by
\begin{align}
\mathbf{x}_{{LR}_{t-1}} \gets \mathbf{x}'_{{LR}_{t-1}} -  {\rho_t}\nabla_{\mathbf{x}_{{LR}_{t}}}\|\mathbf{y} - \mathbf{A}_{LR}\mathbf{x}_{{LR}_{0|t}})\|_2^2.
\label{eq:dps_ti}
\end{align}
In the second stage, we incorporate reconstruction guidance following the standard DPS method given in Eq.~\eqref{eq:ndm core}.
\newpara{\ti-$\mathbf{\Pi}$GDM:} We incorporate pseudoinverse guidance in  both stages. In down-sampled pixel space, pseudoinverse guidance is given by
\begin{equation}
\begin{gathered}
\mathbf{x}_{{LR}_{t-1}}  \gets \mathbf{x}'_{{LR}_{t-1}} -  {\rho_t}\nabla_{\mathbf{x}_{{LR}_{t}}}E(y,\mathbf{A^{\dagger}}_{LR}, \mathbf{x}_{{LR}_{0|t}}),\text{ where,}\\
E(y,\mathbf{A^{\dagger}}_{LR},\mathbf{x}_{{LR}_{0|t}})=\| \mathbf{A^{\dagger}}_{LR}\mathbf{y} - \mathbf{A^{\dagger}}_{LR}\mathbf{A}_{LR}(\mathbf{x}_{{LR}_{0|t}})\|_2^2.
\label{eq:pgd_ti}
\end{gathered}
\end{equation}
The second stage incorporates pseudoinverse guidance following the standard $\Pi$GDM approach given in Eq.~\eqref{eq:pgd_guidance}.
\newpara{\ti-DDNM:} We impose null-space consistency in both stages.
In down-sampled pixel space, $\mathbf{x}_{{LR}_{0|t}}$ is rectified at each step as
\begin{equation}
    \hat{\mathbf{x}}_{{LR}_{0|t}}=\mathbf{A^{\dagger}}_{_{LR}}\mathbf{y} + (\mathbf{I} - \mathbf{A^{\dagger}}_{LR}\mathbf{A}_{LR})\mathbf{x}_{{LR}_{0|t}}.
    \label{eq:ddm_ti}
\end{equation}
The second stage has the usual DDNM null space rectification given by Eq.~\eqref{eq:ndm core}.

In the supplementary material,  we explore text-guided super-resolution using Stable diffusion \cite{rombach2022high}. It is not straightforward to impose null-space consistency on the intermediate estimates in the Stable Diffusion model similar to DDNM, as the diffusion process happens in the latent space. We show in the supplementary that this does not lead to desirable solutions. 
\subsection{CLIP guided Image Super-resolution}
We incorporate CLIP guidance into DDNM-based image super-resolution. For a given image $\mathbf{x}_0$ and text prompt $c$, we define an energy function $E(c,\mathbf{x}_0)$ which measures similarity between the given image and text prompt using CLIP model, through cosine similarity between the CLIP image embeddings and CLIP text embeddings. At each step $t$, we obtain a clean estimate $\mathbf{x}_{0|t}$ using Eq.~\eqref{eq:x0t}, and compute the gradient $\nabla_{\mathbf{x}_t}E(c,\mathbf{x}_{0|t})$.
We rectify $\mathbf{x}_{0|t}$ to satisfy null space consistency using Eq.~\eqref{eq:ndm core} to obtain $\hat{\mathbf{x}}_{0|t}$ and  use it to compute an intermediate estimate of the previous step:
 \begin{equation}
      \hat{\mathbf{x}}_{t-1}\sim p(\mathbf{x}_{t-1}|\mathbf{x}_{t},\hat{\mathbf{x}}_{0|t}).
 \end{equation}
This intermediate previous step is then modified to incorporate CLIP guidance as
\begin{align}
\mathbf{x}_{t-1} \gets  \hat{\mathbf{x}}_{t-1} -  {\rho_t}\nabla_{\mathbf{x}_t}E(c,\mathbf{x}_{0|t}).
\label{eq:clip-guid-prev}
\end{align}
While one could also define an energy function combining  CLIP guidance with reconstruction guidance or pseudo-inverse guidance, we observe that this creates a trade-off between the two objectives of minimizing reconstruction loss and maximizing similarity with text, and does not provide satisfactory results. 

\section{Experimental Evaluation}
 \begin{table*}[t]
     \centering
     \resizebox{\textwidth}{!}{\begin{tabular}{c c c cc cccc}
     \toprule    Dataset&SR&Metric&DPS&DDNM
     &\cite{saharia2022photorealistic}+DDNM&\cite{ramesh2022hierarchical}+DDNM&CLIP guided &\cite{saharia2022photorealistic}+$\Pi$GDM\\
     \midrule
     \multirow{6}{*}{Faces}&\multirow{2}{*}{$8\times$}&LR PSNR(dB)($\uparrow$)& 50.42&75.40
     & 51.68&67.02& 50.16 & 51.08\\
     &&NIQE($\downarrow$)& 5.59& 8.41
     &6.17 & 5.54& 6.12&  6.86\vspace{0.2em}\\
     &\multirow{2}{*}{$16\times$}&LR PSNR(dB)($\uparrow$)& 51.98  & 80.91
     &51.86&66.30 &51.79 &52.02\\
      &&NIQE($\downarrow$)& 5.54& 9.77 
      &5.94& 5.43& 6.38& 6.98\\
      
     \midrule
      \multirow{6}{*}{Nocaps} &\multirow{3}{*}{$8\times$}&LR PSNR(dB)($\uparrow$)&47.01 & 72.94 
      &50.34 & 66.33 &47.86 & 48.75\\
        &&NIQE($\downarrow$)&9.66 &10.27 
        &4.62& 4.88& 5.37&5.10 \\        &&CLIP($\uparrow$)&0.2592&0.2326&0.3102&0.3344&0.2564&0.2811\\
     &\multirow{3}{*}{$16\times$}&LR PSNR (dB)($\uparrow$)&48.07 &78.42 
     & 53.05& 70.01  &50.97 & 49.67\\     
      &&NIQE($\downarrow$)&4.81&13.24 
      & 4.72&5.21 &5.71 & 5.33\\      &&CLIP($\uparrow$)&0.2418&0.2162&0.3037&0.3381&0.2517&0.2788\\
     \bottomrule
     \end{tabular}}
     \caption{Quantitative evaluation of baselines DPS\cite{chung2022diffusion},\cite{wang2022zero}, and Imagen\cite{saharia2022photorealistic}+DDNM, \cite{ramesh2022hierarchical}, CLIP guided DDNM, Imagen\cite{saharia2022photorealistic}+$\Pi$GDM.}
     \label{tab:quant}
     \end{table*}
We perform experiments on extreme super-resolution of images with large super-resolution factors $\times8$, $\times16$, as this problem is severely ill-posed and allows exploration of a larger solution space, and is therefore an ideal setting to test our method on exploring diverse solutions. In contrast, input imposes stronger constraints on the solutions for super-resolution at smaller scale factors, limiting their diversity and explorability. 
 We generate low-resolution images using bicubic downsampling and compute the pseudoinverse operator $\mathbf{A}^\dagger$ using SVD following  \cite{kawar2022denoising,wang2022zero}. To evaluate consistency between the generated result and the input text prompt, we use CLIP score \cite{radford2021learning} using the ViT-B/16 CLIP model. For super resolutions with large factors, PSNR/SSIM which measure consistency with ground truth are not effective metrics to measure reconstruction performance, as multiple solutions can lead to the same low-resolution image. We instead evaluate consistency by calculating PSNR between the input LR image and the downsampled version of the solution,  as also used in recent challenges for learning super-resolution space \cite{gu2022ntire,lugmayr2021ntire}. We evaluate the reconstruction quality in terms of NIQE score \cite{mittal2012making}. We also conduct a user study to evaluate how the users rate the plausibility of reconstruction and semantic consistency with the text prompt. 

 We perform experiments on face images from CelebA-HQ dataset \cite{karrasprogressiveCelebAHQ} (a subset of 200 images) and open domain images with captions from NoCaps dataset~\cite{agrawal2019nocaps} (a subset of 100 images). For face image super-resolution, we manually provide different text prompts with varying personal attributes such as age, gender, smile, glasses, and curly hair.  
 All our experiments are performed for an output resolution of $256\times256$. For text-guided restoration with \ti models, we use the open-source versions of Imagen Deep-Floyd IF~\cite{deepfloyd}, unCLIP  Karlo-unCLIP~\cite{kakaobrain2022karlo-v1-alpha}, and Stable Diffusion v1.4 \cite{rombach2022high}. We describe the detailed settings and hyper-parameters for each method in the supplementary material. We will make our code publicly available.
\begin{figure}[htb]
\centering
\scriptsize
  \includegraphics[width=0.24\linewidth]{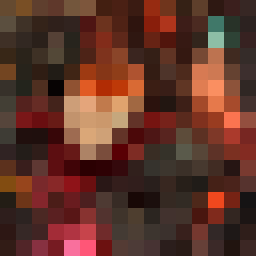}
\includegraphics[width=0.24\linewidth]{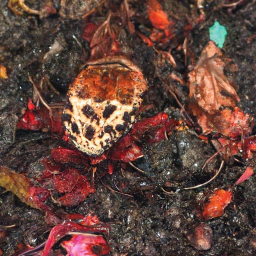}
\includegraphics[width=0.24\linewidth]{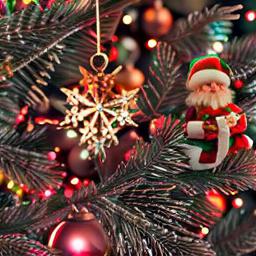}
\includegraphics[width=0.24\linewidth]{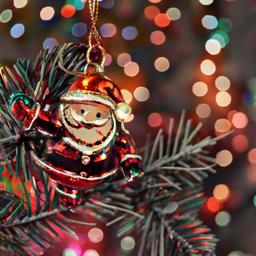}
\vspace{0.15em}\\
The Santa ornament hangs from a Christmas tree branch amongst the colorful bright lights.\\
\includegraphics[width=0.24\linewidth]{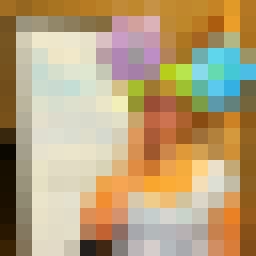}
\includegraphics[width=0.24\linewidth]{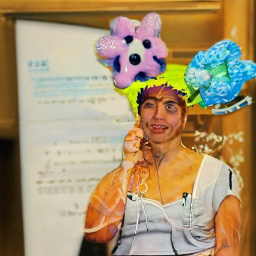}
\includegraphics[width=0.24\linewidth]{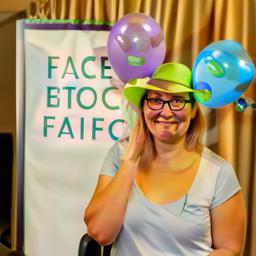}
\includegraphics[width=0.24\linewidth]{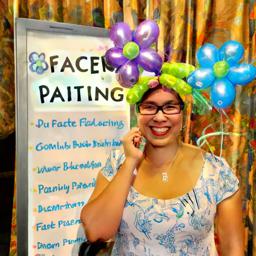}
\vspace{0.15em}\\
A woman wearing glasses and smiling has a green balloon hat with one purple balloon flower and one blue balloon flower in front of a whiteboard that says face painting\\
\includegraphics[width=0.24\linewidth]{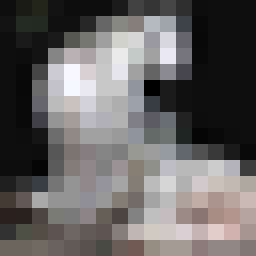}
\includegraphics[width=0.24\linewidth]{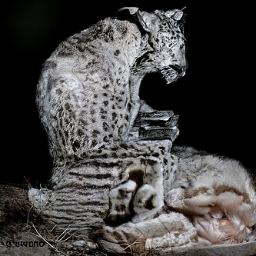}
\includegraphics[width=0.24\linewidth]{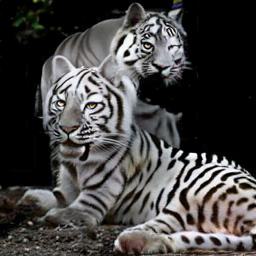}
\includegraphics[width=0.24\linewidth]{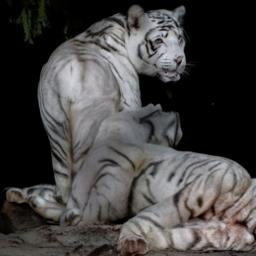}
\vspace{0.15em}\\
Two white tigers, one laying down, the other standing.
\\
\phantom{AAAAAA}\footnotesize{LR}\phantom{AAAAA}\footnotesize{DPS}\cite{chung2022diffusion}\phantom{A}{\hspace*{0.05\linewidth}}\text{\footnotesize{\cite{ramesh2022hierarchical}+DDNM} }\phantom{A}{\hspace*{0.05\linewidth}}\text{\footnotesize{\cite{saharia2022photorealistic}+DDNM} }
 \\
\caption{Visual comparison of $16\times$ SR on open domain images.\label{fig:nocaps_sr}}
\end{figure}
\begin{figure*}[htb]
\centering
\scriptsize
  \includegraphics[width=0.151\linewidth]{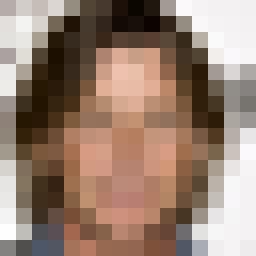}
\includegraphics[width=0.151\linewidth]{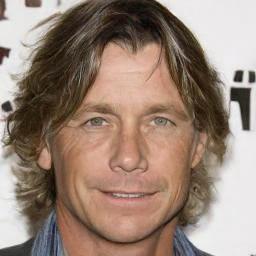}
\includegraphics[width=0.151\linewidth]{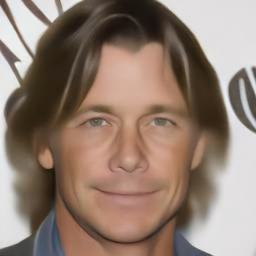}
\includegraphics[width=0.151\linewidth]{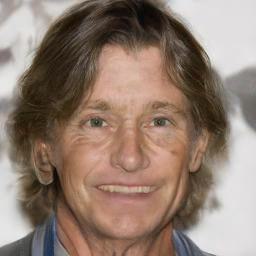}
\includegraphics[width=0.151\linewidth]{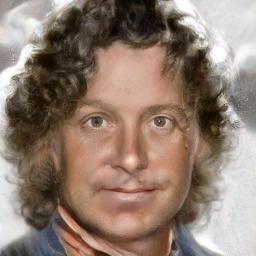}
\includegraphics[width=0.151\linewidth]{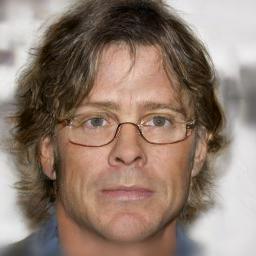}\\
\phantom{AAA}\footnotesize{LR}\phantom{AAAAAAAAAA}\footnotesize{DPS}\cite{chung2022diffusion}\phantom{AAAAAAAAAAA}DDNM\cite{wang2022zero}\phantom{AAA}$\xleftarrow{\hspace*{0.13\linewidth}}\text{CLIP guidance+DDNM} \xrightarrow{\hspace*{0.13\linewidth}}$\\
\includegraphics[width=0.151\linewidth]{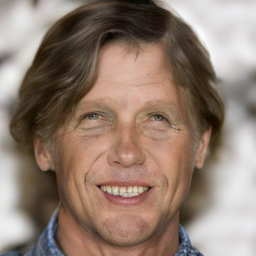}
\includegraphics[width=0.151\linewidth]{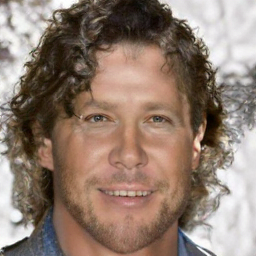}
\includegraphics[width=0.151\linewidth]{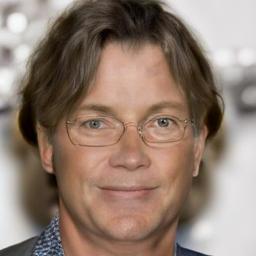}
\includegraphics[width=0.151\linewidth]{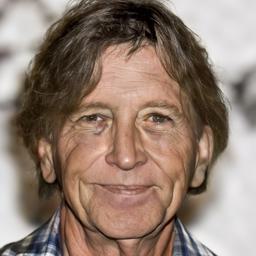}
\includegraphics[width=0.151\linewidth]{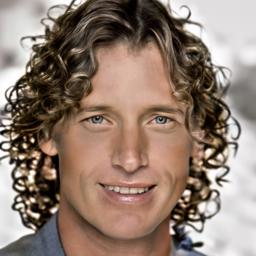}
\includegraphics[width=0.151\linewidth]{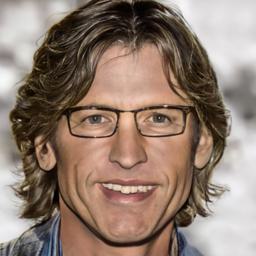}
 \\$\xleftarrow{\hspace*{0.17\linewidth}}\text{\cite{ramesh2022hierarchical}+DDNM} \xrightarrow{\hspace*{0.17\linewidth}}\xleftarrow{\hspace*{0.17\linewidth}}\text{\cite{saharia2022photorealistic}+DDNM} \xrightarrow{\hspace*{0.17\linewidth}}$
\caption{Exploring solution for $16\times$ SR on a face image for the prompts: `Elderly smiling man', ` Man with curly hair', `Man with glasses',\label{fig:142}}
\end{figure*}

      We first compare the performance of proposed baselines against existing restoration methods,  vanilla DDNM and DPS using unconditional diffusion models trained on CelebA-HQ faces~\cite{karrasprogressiveCelebAHQ} and Imagenet \cite{deng2009imagenet}.        
      For this experiment, we provide a neutral prompt for text-based methods `a high-resolution photograph of a face' for face images, and utilize images and captions from the NoCaps dataset for open domain images.    The results are summarized in Tab.~\ref{tab:quant}. Among the zero-shot diffusion-based image restoration methods,  DDNM achieves the best LR PSNR owing to exact consistency imposed by projection operation, where DPS achieves visually sharper results, which is reflected in the lower NIQE scores. The text-based reconstruction baselines from Sec.~\ref{sec:methodology}  achieve comparable performance to specialized models trained on faces on neutral prompts.  Further, on open-domain images from the NoCaps dataset, we observe significantly better image quality in terms of NIQE score, and semantic matching as measured by CLIP score in comparison to DDNM and DPS using diffusion model trained on Imagenet. Further, we note that using CLIP guidance along with DDNM significantly improves both semantic matching with text as well as NIQE, with a reduction in LR consistency when compared to vanilla DDNM which yields rather blurred results. All the methods still maintain a good consistency with the low-resolution measurement, with LR PSNR $>$ 45 dB.        
      In addition to the baselines in Tab.~\ref{tab:quant}, we also experimented with the baselines of \cite{choi2021ilvr, jo2020investigating}, we did not include these results as these methods do not achieve the desired level of LR consistency. Among these \cite{jo2020investigating} is trained for the task of extreme super-resolution, yet, we find that it cannot handle very low-resolution inputs well.
      
Fig.~\ref{fig:nocaps_sr} qualitatively compares the super-resolution performance of vanilla DPS with \ti-DDNM approaches using unCLIP and Imagen.  On this test set, DDNM using an unconditional diffusion model trained on Imagenet produces blurry results. While DPS recovers sharp images in where the image content is simpler, Figs.~\ref{fig:nocaps_sr} and~\ref{fig:teaser} show that it struggles with complex scene content. On the other hand, the use of powerful \ti models in zero-shot restoration can recover data consistent solutions matching complex text prompts.
\newpara{\ti$-\mathbf{\Pi}$GDM and \ti-DPS}  Among the \ti model based approaches, Imagen~\cite{saharia2022photorealistic}+$\Pi$GDM has the least CLIP score in Tab.~\ref{tab:quant}. This is because this method is evaluated without classifier-free guidance (CFG)~\cite{ho2022classifier}, whereas DDNM-based methods included CFG. It is known that including CFG in \ti models improves adherence to text. However, we find that it conflicts with gradient-based measurement guidance, reducing data consistency in \ti$-\mathbf{\Pi}$GDM and \ti-DPS. The use of CFG requires a lower stepsize parameter, and as seen in Tab.~\ref{tab:pigdm}, the use of the same number of diffusion steps drastically decreases LR-PSNR while improving CLIP score on the NoCaps dataset.  We investigate this further by evaluating \cite{saharia2022photorealistic}+$\Pi$GDM with varying numbers of reverse diffusion steps on a subset of 25 CelebAHQ images for the prompt `a photograph of a woman with curly hair'. While increasing the number of diffusion steps improves LR PSNR, it also reduces text adherence. We see that improving LR PSNR always does not lead to desired results in Fig.~\ref{fig:pgdm}.  In the supplementary material Sec.~\ref{sec:t2i_gradient}, we include a detailed study and include results of similar experiments with Imagen \cite{saharia2022photorealistic}+DPS. We observe that Imagen$-$DPS does not provide the desired level of  LR consistency even without classifier-free guidance. When classifier-free guidance is included, text adherence improves, with a significant drop in LR  PSNR. 
\paragraph{Exploring solutions through text }
 Figs.~\ref{fig:teaser} and \ref{fig:142} provide qualitative comparisons of the proposed text-based baselines with DPS \cite{chung2022diffusion} and \cite{wang2022zero} on face images. While vanilla DPS and  DDNM using a diffusion model trained on faces achieve realistic and data-consistent solutions, they offer little scope for exploration and produce solutions with limited diversity.  On the other hand, the proposed baselines can recover images with great diversity in attributes such as curly hair, glasses, expression, and age. As the ill-posedness of the recovery problem becomes more severe at high SR factors $(\times32)$, it is possible to recover a wide variety of outputs with challenging attributes in age, race, and appearance. We show more examples in the supplementary material. 

\begin{table}[]
    \centering
\resizebox{0.49\textwidth}{!}{\begin{tabular}{c|c|c|c|c|c}
\toprule
&Step size& Steps&CFG&LR PSNR (dB)&CLIP score\\
\midrule
 \multirow{2}{*}{Nocaps} &1.0 &  100&\ding{55}&49.67&0.2788\\
  & 0.5  & 100&\ding{51}&24.31&0.2923\\
   \midrule
 \multirow{4}{*}{Faces}  &1.0&100&\ding{55}&48.705&0.2560\\
   &0.5&100&\ding{51}&30.675&0.3011\\
   &0.5&200&\ding{51}&40.025&0.2742\\
   &0.5&300&\ding{51}&42.431&0.2773\\
   &0.5&500&\ding{51}&43.093&0.2695\\
   \bottomrule
\end{tabular}}
    \caption{Effect of CFG, step size, and number of steps on consistency and text adherence in \ti-$\Pi$GDM .\label{tab:pigdm}}
\end{table}
\begin{figure}[htb]
\centering
\scriptsize
\includegraphics[width=0.19\linewidth]{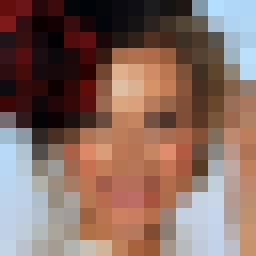}
   \includegraphics[width=0.19\linewidth]{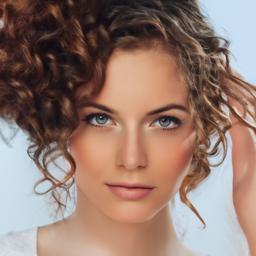}
\includegraphics[width=0.19\linewidth]{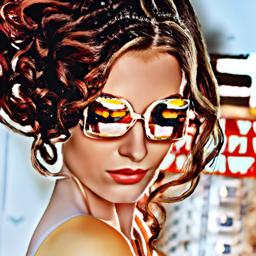}
\includegraphics[width=0.19\linewidth]{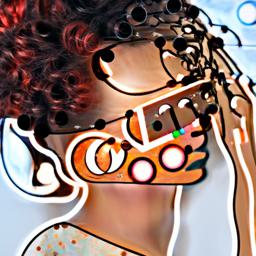}
\includegraphics[width=0.19\linewidth]{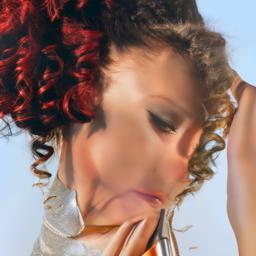}
\vspace{0.15em}\\
LR \qquad\qquad 28.34dB\qquad\qquad28.09dB\qquad\qquad34.43dB\qquad\qquad41.56dB\\
\includegraphics[width=0.19\linewidth]{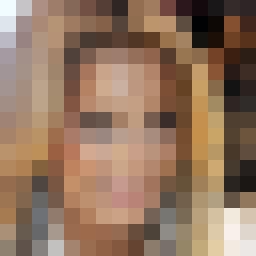}
   \includegraphics[width=0.19\linewidth]{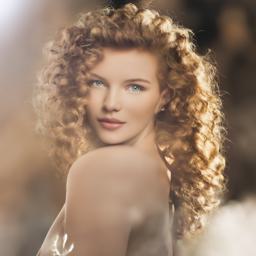}
\includegraphics[width=0.19\linewidth]{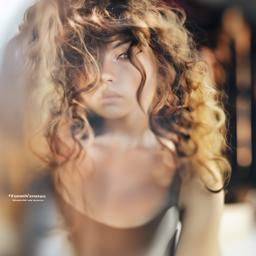}
\includegraphics[width=0.19\linewidth]{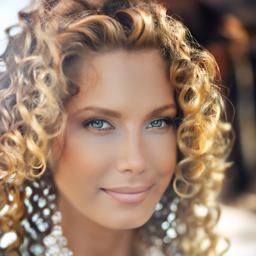}
\includegraphics[width=0.19\linewidth]{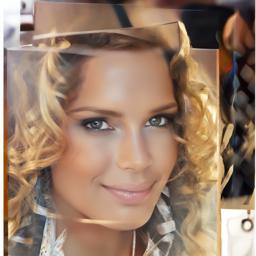}
\vspace{0.15em}\\
LR \qquad\qquad27.25dB\qquad\qquad41.28dB\qquad\qquad41.81dB\qquad\qquad41.59dB\\
\includegraphics[width=0.19\linewidth]{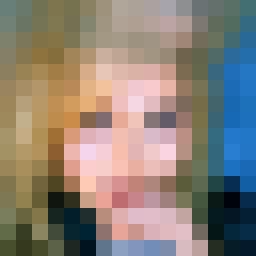}
   \includegraphics[width=0.19\linewidth]{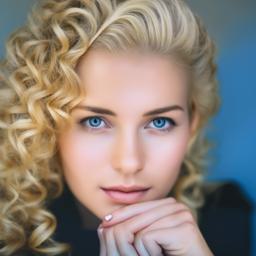}
\includegraphics[width=0.19\linewidth]{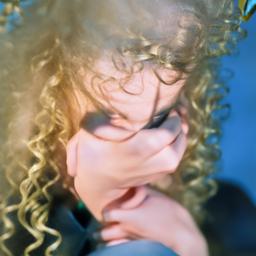}
\includegraphics[width=0.19\linewidth]{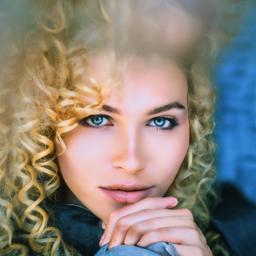}
\includegraphics[width=0.19\linewidth]{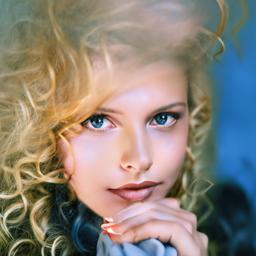}
\vspace{0.15em}\\
LR \qquad\qquad20.09dB\qquad\qquad43.5dB\qquad\qquad44.0dB\qquad\qquad44.2dB\\
\qquad\qquad\quad100steps\qquad\quad200steps\qquad\quad300steps\qquad\qquad500steps\\

\caption{Effect of classifier-free guidance and stepsize in Imagen-$\Pi$GDM.\label{fig:pgdm}}
\end{figure}
\comment{\begin{table}[]
    \centering
\resizebox{0.499\textwidth}{!}{\begin{tabular}{c|cc|cc|cc}
\toprule
& \multicolumn{2}{c|}{Imagen-DDNM}&\multicolumn{2}{c|}{unCLIP-DDNM}&\multicolumn{2}{c}{CLIP guidance}\\
&LR PSNR&CLIP&LR PSNR&CLIP&LR PSNR&CLIP\\
\midrule
  Woman, curly hair &\\
   elderly woman &\\
   Smiling man&\\
   Man with glasses&\\
   \bottomrule
\end{tabular}}
    \caption{Quantitative evaluation on exploring solutions on Celeb A HQ $16\times$ SR\label{tab:explore}}
\end{table}}
\begin{figure}[t]
  \begin{center}
  \includegraphics[width=0.245\linewidth]{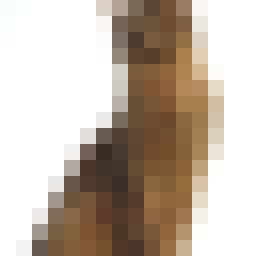}
  \hspace{-1pt}\includegraphics[width=0.245\linewidth]{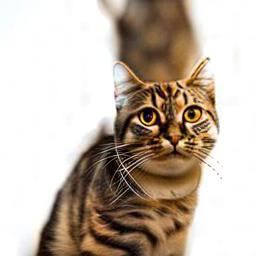}
  \hspace{-2pt}\includegraphics[width=0.245\linewidth]{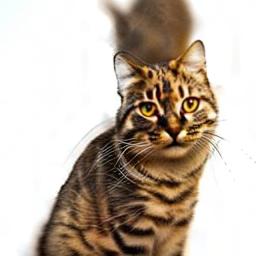}
  \hspace{-2pt}\includegraphics[width=0.245\linewidth]{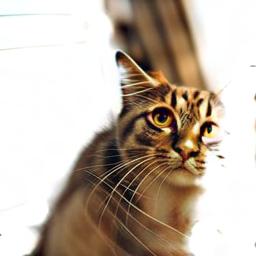}\\
   \includegraphics[width=0.245\linewidth]{images/sr_embd_avg/16xLR00032.jpg}
  \hspace{-1pt}\includegraphics[width=0.245\linewidth]{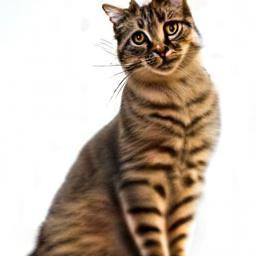}
  \hspace{-2pt}\includegraphics[width=0.245\linewidth]{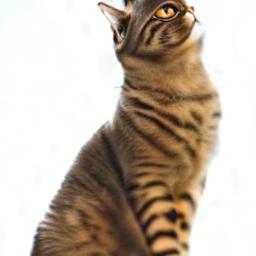}
  \hspace{-2pt}\includegraphics[width=0.245\linewidth]{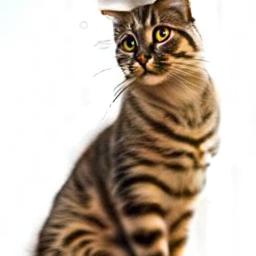}\\
  \caption{$\times16$ SR results with (bottom) and without (top) averaging trick with $\lambda$$=$$0.4$, and the text prompt `a high-res photo of a cat'.}\label{fig:icml_avg_trick}
  \end{center}
\end{figure} 
\paragraph{Limitation of \ti-DDNM}
While \ti-DDNM methods do not face a trade-off between text adherence and measurement consistency, they can still result in unrealistic images that are both consistent with the text as well as measurement. This is because any image hallucinated by the \ti model can still be made consistent with the measurement through null-space component rectification. We sometimes encounter this problem in unCLIP\cite{ramesh2022hierarchical}-DDNM, when image embedding imagined by prior does not structurally align with the observation. This can be mitigated to an extent by modifying the image embedding which conditions the diffusion model. We introduce an embeddings averaging trick by considering a convex combination of embedding provided by the prior, and CLIP image embedding of the pseudo-inverse solution. The extent of this averaging can be controlled by an additional hyperparameter $\lambda$ which determines the weight of pseudoinverse embedding. This embeddings averaging trick can improve the structural consistency of the solution with the input observation, as seen in Fig.~\ref{fig:icml_avg_trick}.
\paragraph{User Study}
We performed a user study to evaluate the realism and semantic matching with text prompts on the results of Imagen-DDNM, unCLIP DDNM, and CLIP-guided DDNM using an online survey platform. The survey included 50 reconstructions for each method for the task of $\times16$ super-resolution along with the corresponding text prompts. The LR images were generated using 30 face images from CelebA HQ, and 20 open-domain images from the NoCaps dataset. The users were asked to evaluate separately whether each reconstruction semantically matches with input text prompt and whether the solution appears photo-realistic. The results of this survey are found in Tab.~\ref{tab:survey}. Both Imagen DDNM and unCLIP DDNM score better in terms of user preference in comparison to CLIP-guided recovery. This is to be expected, as both \cite{kakaobrain2022karlo-v1-alpha} and \cite{deepfloyd} are powerful \ti models trained over webscale data. Among the three methods, Imagen DDNM has the best user preference in terms of both semantic matching with text as well as perceived realism of the recovered solution. This human evaluation is in contrast with the quantitative evaluation in Tab.~\ref{tab:quant}  where unCLIP achieves higher semantic matching with text in terms of CLIP score. As unCLIP was trained to invert CLIP embeddings, it possibly produces images with higher CLIP scores.
\begin{table}[]
    \centering
\resizebox{0.499\textwidth}{!}{\begin{tabular}{c|cc|cc|cc}
\toprule
& \multicolumn{2}{c|}{Imagen-DDNM}&\multicolumn{2}{c|}{unCLIP-DDNM}&\multicolumn{2}{c}{CLIP guidance}\\
&Faces&nocaps&Faces&nocaps&Faces&nocaps\\
\midrule
  Text-Similarity &90.89\% &92.88\%&73.20\%&54.42\% &60.64\%&25.38\%\\
   Photo-realism & 69.35\%&83.07\%&30.89\%&35.57\%&30.77\%&10.38\%\\
   \bottomrule
\end{tabular}}
    \caption{Results of user survey on text guided super-resolution\label{tab:survey}}
\end{table}

\section{Related Work}
\textbf{Diverse Solutions to Image Super-resolution}~ Deep networks have become popular tool for image super-resolution in the past decade\cite{wang2018esrgan,ledig2017photo,menon2020pulse,choi2021ilvr}, where many state of the art methods \cite{jo2020investigating,chan2021glean,wang2022panini,wang2021real} employ supervised training to recover a single solution. Deep learning based solutions which allow sampling multiple solutions to the ill-posed SR problem also exist, \cite{bahat2020explorable,lugmayr2020srflow,buhler2020deepsee,kawar2022denoising,chung2022come,wang2022zero,song2023pseudoinverseguided,lugmayr2020srflow,Jo_2021_CVPR, choi2021ilvr,menon2020pulse,explore_solution_gan_latent}. These methods utilize conditional or unconditional generative models to sample solutions. Among these conditional generative model based approaches \cite{bahat2020explorable,lugmayr2020srflow,saharia2021image,li2022srdiff,peng2020generating,jo2021tackling,lugmayr2021ntire, Lugmayr_2022_CVPR} are  trained for the specific super-resolution task. Zero-shot approaches\cite{menon2020pulse,kawar2022denoising,chung2022come,wang2022zero,song2023pseudoinverseguided} on the other hand, utilize image generative models directly for image recovery. 
\paragraph{Explorable Image Super-resolution}  
A few prior works attempt to explore solutions space
 using graphical inputs \cite{bahat2020explorable} or semantic maps \cite{buhler2020deepsee}. However, they are still restricted to specific classes e.g. faces, or trained for specific degradation, e.g. specific super-resolution factors.  A recent work \cite{ma2022rethinking} also combines text features into super-resolution network architectures using attention, and trains separate models for text-guided image super-resolution in an end-to-end manner for each dataset and super-resolution factor.  Yet, this approach cannot handle open domain images and arbitrary super-resolution factors. To the best of our knowledge, there is no existing method that allows zero-shot exploration of solutions space for different restoration tasks on open-domain images through text.
\paragraph{Diffusion Models for Image Super-resolution} 
One could utilize diffusion models for image super-resolution and other restoration tasks either by training a conditional diffusion model for specific tasks \cite{saharia2021image,li2022srdiff,Whang_2022_CVPR,zhao2023towards},  or by leveraging diffusion models for zero-shot image recovery \cite{jalal2021robust,kadkhodaie2021stochastic,choi2021ilvr,chung2022come,chung2022improving,kawar2021snips,wang2022zero,lugmayr2022repaint,kawar2022denoising,Nair_2023_ICCV}. We are concerned with the latter variety, which exploit the knowledge of degradation operator to modify the sampling process. Earlier works \cite{jalal2021robust,kadkhodaie2021stochastic} adopt Langevin dynamics for linear inverse problems and incorporate measurement guidance through the gradient of the least-squares data fidelity term. \cite{choi2021ilvr,chung2022come} alternate between a standard reverse diffusion step and a projection step promoting measurement consistency. Recent works utilize an estimate of clean sample at each reverse step to modify the sampling process via a consistency enforcing projection operation \cite{wang2022zero,lugmayr2022repaint,kawar2022denoising} or through guidance through the gradient of the least-squares data fidelity term \cite{chung2022diffusion}, or least squares measurement guidance  \cite{song2023pseudoinverseguided}
or both \cite{chung2022improving}. While projection-based approaches are faster and do not need to backpropagate through diffusion model weights,  they are restricted to inverse problems where a pseudo-inverse or its approximation can be computed. On the other hand, gradient-based measurement guidance can be applied for any inverse problems, or even arbitrary guidance \cite{universal2023,yu2023freedom}, yet it is more expensive as it requires back-propagation through the diffusion model weights at each iteration.  More recently, \cite{mardani2023variational} adopt diffusion models in a regularization by denoising (RED) framework, and \cite{zhu2023denoising} demonstrate their utility for plug-and-play image restoration as an effective alternative to the standard Gaussian denoisers. 
\paragraph{\ti Generative Models}
Starting from  \cite{mansimov2015generating} many works proposed different methods to generate images from text prompts. Initial works trained RNNs \cite{mansimov2015generating} and GANs \cite{reed2016generative, zhang2017stackgan,xu2018attngan, zhang2017stackgan,zhang2018stackgan++,dmgan, li2019controllable, zhang2021cross, zhu2022label} using attention on smaller captioned image datasets~\cite{Wah2011TheCB,Lin2014MicrosoftCC}.  Recent developments in image generation~\cite{dhariwal2021diffusion,nichol2021improved,esser2021taming}, contrastive learning\cite{Chen2020ASF} and large-scale training on massive internet-scale datasets of paired text prompts and images~\cite{Schuhmann2022LAION5BAO,schuhmann2021laion} accelerated research in vision-language learning \cite{radford2021learning,jia2021scaling,ramesh2021zero, nichol2021glide, ramesh2022hierarchical,saharia2022photorealistic}.  Many recent works train text-to-image (T2I) models directly on large scale datasets using autoregressive transformers \cite{ramesh2021zero,ding2021cogview,gafni2022make} or diffusion based models\cite{nichol2021glide,ramesh2022hierarchical,saharia2022photorealistic}.
Some of these T2I models perform diffusion in a low dimensional latent space \cite{gu2021vector,tang2022improved,rombach2022high,bond2021unleashing,esser2021imagebart, Hu_2022_CVPR}, or in a down-sampled pixel space \cite{ramesh2022hierarchical,saharia2022photorealistic} for computational efficiency. 
An alternative paradigm is employed by \cite{galatolo2021,crowson2022vqgan,clip_guided_diffusion,liu2021fusedream, liu2021more,couairon2022flexit,paiss2022no}  using text-image encoder CLIP~\cite{radford2021learning}  approaches to guide pretrained generative  models \cite{esser2021taming, dhariwal2021diffusion} towards the input text prompt.

\section{Discussion, Limitations, and Conclusions}
In this paper, we introduced the challenging task of zero-shot open-domain extreme super-resolution for different scale factors guided by text prompts. We explored two approaches to deal with this challenge- utilizing pretrained diffusion-based \ti models for zero-shot recovery and by guiding an image diffusion model with CLIP for zero-shot diffusion-based restoration. We showed that the proposed methods improve adherence to input text prompts while maintaining consistency with the observation. We demonstrated significantly improved diversity in solutions using the proposed methods. Among these methods, CLIP guidance is naturally outperformed by the more powerful \ti diffusion model-based methods. Further, we found that gradient-based reconstruction guidance could be in trade-off with text adherence.  Moreover, the generated results are not always realistic, and it can require several attempts to realize the desired output, as also observed in text condition image generation \cite{karthik2023dont}. 
It must be noted that not every text prompt is meaningful for every observation. When certain patterns or objects indicated by text cannot be present in the image, the corresponding objects or patterns cannot be recovered without severe artifacts or unrealistic images. In this case, it is not the failure of the approach or the model, rather it can help the users determine the plausibility of a solution. In view of this, the evaluation of text-guided restoration is highly subjective, and any quantitative evaluation in terms of image quality metrics is meaningful only when the input text prompts are well aligned and plausible for the given degraded measurement. 
The performance of all the proposed methods depends on and is limited by the generative capabilities of the pre-trained generative model.  The method inherits the biases of the data used to train the \ti model. Finally, our work opens up a promising direction of developing efficient user-guided tools for text-based exploration of image recovery.

{
    \small
    \bibliographystyle{ieeenat_fullname}
    
}

\clearpage
\maketitlesupplementary
This supplementary material is organized as follows:
\cref{sec:algos} provides the algorithms for text based super-resolution introduced in \cref{sec:methodology}.
We conduct further experiments demonstrating the trade-off between text adherence and LR PSNR in gradient-based guidance methods in \cref{sec:t2i_gradient}, and show the results of our experiments with other approaches.
We provide detailed experimental settings in \cref{{sec:settings}} and additional qualitative results in \cref{sec:res_supple}. 
\section{Algorithms for Text based Super-resolution}\label{sec:algos}
We provide the algorithms for text-based super-resolution methods described in \cref{sec:methodology}. In the following $p_1$ and $p_2$ denote reverse diffusion processes in the text-conditioned downsampled stage and the subsequent super-resolution stage in \ti diffusion models. For Imagen \cite{saharia2022photorealistic}, $\mathbf{c}_1$ corresponds to text embeddings from text encoder, and $\mathbf{c}_2$ contains $\mathbf{x}_{LR}$, in addition to text embeddings. For unCLIP \cite{ramesh2022hierarchical} $\mathbf{c}_1$ corresponds to a combination of  CLIP image embeddings produced by the prior model and text embeddings, and $\mathbf{c}_2$ is the output $\mathbf{x}_{LR}$ of the first stage.
\begin{algorithm}[h]
    \begin{algorithmic}[T]
        \State $\mathbf{x}_{{LR}_T}\sim\mathcal{N}(\mathbf{0},\mathbf{I})$
        \For{$t = T, ..., 1$}
            \State $\mathbf{x}_{{LR}_{0|t}} = \frac{1}{\sqrt{\Bar{\alpha}_{t}}}	\left( \mathbf{x}_{LR_t} - \boldsymbol{\epsilon_\theta}(\mathbf{x}_{LR_t},t|\mathbf{c}_1)\sqrt{1-\Bar{\alpha}_{t}} \right)$
            \State $ \hat{\mathbf{x}}_{{LR}_{0|t}}=\mathbf{A^{\dagger}}_{_{LR}}\mathbf{y} + (\mathbf{I} - \mathbf{A^{\dagger}}_{LR}\mathbf{A}_{LR})\mathbf{x}_{{LR}_{0|t}}$
            \State $\mathbf{x}_{{LR}_{t-1}}\sim p_1(\mathbf{x}_{{LR}_{t-1}}|\mathbf{x}_{{{LR}}_t},\hat{\mathbf{x}}_{{LR}_{0|t}})$
        \EndFor
        \State $\mathbf{x}_{LR}\leftarrow\mathbf{x}_{LR_0}$ 
         \State $\mathbf{x}_{T}\sim\mathcal{N}(\mathbf{0},\mathbf{I})$
        \For{$t = T, ..., 1$}
            \State $\mathbf{x}_{0|t} = \frac{1}{\sqrt{\Bar{\alpha}}_{t}}\left( \mathbf{x}_{t} - \boldsymbol{\zeta_\theta}(\mathbf{x}_{t},t|\mathbf{c}_2)\sqrt{1-\Bar{\alpha}_{t}} \right)$
            \State $ \hat{\mathbf{x}}_{0|t} = \mathbf{A}^{\dagger}\mathbf{y} + (\mathbf{I} - \mathbf{A}^{\dagger}\mathbf{A})\mathbf{x}_{0|t}$
            \State $\mathbf{x}_{t-1}\sim p_2(\mathbf{x}_{t-1}|\mathbf{x}_{t},\hat{\mathbf{x}}_{0|t})$
        \EndFor
        \State \textbf{return} $\mathbf{x}_{0}$        
    \end{algorithmic}
    \caption{\ti-DDNM sampling process }
    \label{alg:ndm_unclip}
    \end{algorithm}
\begin{algorithm}[h]
    \begin{algorithmic}[T]
        \State $\mathbf{x}_{{LR}_T}\sim\mathcal{N}(\mathbf{0},\mathbf{I})$
        \For{$t = T, ..., 1$}
            \State $\mathbf{x}_{{LR}_{0|t}} = \frac{1}{\sqrt{\Bar{\alpha}_{t}}}	\left( \mathbf{x}_{LR_t} - \boldsymbol{\epsilon_\theta}(\mathbf{x}_{LR_t},t|\mathbf{c}_1)\sqrt{1-\Bar{\alpha}_{t}} \right)$
            \State $\mathbf{x}'_{{LR}_{t-1}}\sim p_1(\mathbf{x}_{{LR}_{t-1}}|\mathbf{x}_{{{LR}}_t},{\mathbf{x}}_{{LR}_{0|t}})$
            \State $\mathbf{x}_{{LR}_{t-1}} = \mathbf{x}'_{{LR}_{t-1}} -  {\rho_t}\nabla_{\mathbf{x}_{{LR}_{t}}}\|\mathbf{y} - \mathbf{A}_{LR}\mathbf{x}_{{LR}_{0|t}})\|_2^2$
        \EndFor
        \State $\mathbf{x}_{LR}\leftarrow\mathbf{x}_{LR_0}$ 
         \State $\mathbf{x}_{T}\sim\mathcal{N}(\mathbf{0},\mathbf{I})$
        \For{$t = T, ..., 1$}
            \State $\mathbf{x}_{0|t} = \frac{1}{\sqrt{\Bar{\alpha}}_{t}}\left( \mathbf{x}_{t} - \boldsymbol{\zeta_\theta}(\mathbf{x}_{t},t|\mathbf{c}_2)\sqrt{1-\Bar{\alpha}_{t}} \right)$
            \State $\mathbf{x}'_{t-1}\sim p_2(\mathbf{x}_{t-1}|\mathbf{x}_{t},{\mathbf{x}}_{0|t})$
            \State $\mathbf{x}_{t-1} = \mathbf{x}'_{t-1} -  {\rho_t}\nabla_{\mathbf{x}_t}\|\mathbf{y} - \mathbf{A}(\mathbf{x}_{0|t})\|_2^2$
        \EndFor
        \State \textbf{return} $\mathbf{x}_{0}$        
    \end{algorithmic}
    \caption{\ti-DPS sampling process }
    \label{alg:dps_t2i}
    \end{algorithm}    
\begin{algorithm}[h]
    \begin{algorithmic}[T]
        \State $\mathbf{x}_{{LR}_T}\sim\mathcal{N}(\mathbf{0},\mathbf{I})$
        \For{$t = T, ..., 1$}
            \State $\mathbf{x}_{{LR}_{0|t}} = \frac{1}{\sqrt{\Bar{\alpha}_{t}}}	\left( \mathbf{x}_{LR_t} - \boldsymbol{\epsilon_\theta}(\mathbf{x}_{LR_t},t|\mathbf{c}_1)\sqrt{1-\Bar{\alpha}_{t}} \right)$
            \State $\mathbf{x}'_{{LR}_{t-1}}\sim p_1(\mathbf{x}_{{LR}_{t-1}}|\mathbf{x}_{{{LR}}_t},{\mathbf{x}}_{{LR}_{0|t}})$
            \State $\mathbf{x}_{{LR}_{t-1}} = \mathbf{x}'_{{LR}_{t-1}} -  {\rho_t}\nabla_{\mathbf{x}_{{LR}_{t}}} \| \mathbf{A^{\dagger}}_{LR}\mathbf{y} - \mathbf{A^{\dagger}}_{LR}\mathbf{A}_{LR}(\mathbf{x}_{{LR}_{0|t}})\|_2^2$         
        \EndFor
        \State $\mathbf{x}_{LR}\leftarrow\mathbf{x}_{LR_0}$ 
         \State $\mathbf{x}_{T}\sim\mathcal{N}(\mathbf{0},\mathbf{I})$
        \For{$t = T, ..., 1$}
            \State $\mathbf{x}_{0|t} = \frac{1}{\sqrt{\Bar{\alpha}}_{t}}\left( \mathbf{x}_{t} - \boldsymbol{\zeta_\theta}(\mathbf{x}_{t},t|\mathbf{c}_2)\sqrt{1-\Bar{\alpha}_{t}} \right)$
            \State $\mathbf{x}'_{t-1}\sim p_2(\mathbf{x}_{t-1}|\mathbf{x}_{t},{\mathbf{x}}_{0|t})$
            \State $\mathbf{x}_{t-1} = \mathbf{x}'_{t-1} -  {\rho_t}\nabla_{\mathbf{x}_t}\| \mathbf{A^{\dagger}}\mathbf{y} -  \mathbf{A^{\dagger}}\mathbf{A}(\mathbf{x}_{0|t})\|_2^2$
        \EndFor
        \State \textbf{return} $\mathbf{x}_{0}$        
    \end{algorithmic}
    \caption{\ti-$\Pi$GDM sampling process }
    \label{alg:pigdm_t2i}
    \end{algorithm}        
\begin{algorithm}[h]
    \begin{algorithmic}[T]
         \State $\mathbf{x}_{T}\sim\mathcal{N}(\mathbf{0},\mathbf{I})$
        \For{$t = T, ..., 1$}
            \State $\mathbf{x}_{0|t} = \frac{1}{\sqrt{\Bar{\alpha}}_{t}}\left( \mathbf{x}_{t} -\boldsymbol{\epsilon_\theta}(\mathbf{x}_{t},t)\sqrt{1-\Bar{\alpha}_{t}} \right)$
            \State $ \hat{\mathbf{x}}_{0|t} = \mathbf{A}^{\dagger}\mathbf{y} + (\mathbf{I} - \mathbf{A}^{\dagger}\mathbf{A})\mathbf{x}_{0|t}$
            \State $ \hat{\mathbf{x}}_{t-1}\sim p(\mathbf{x}_{t-1}|\mathbf{x}_{t},\hat{\mathbf{x}}_{0|t})$
            \State $\mathbf{x}_{t-1} \gets  \hat{\mathbf{x}}_{t-1} -  {\rho_t}\nabla_{\mathbf{x}_t}E(c,\mathbf{x}_{0|t})$
        \EndFor
        \State \textbf{return} $\mathbf{x}_{0}$        
    \end{algorithmic}
    \caption{CLIP guidance with DDNM}
    \label{alg:clip_ddnm}
    \end{algorithm}   
\section{Further Analysis of Text based SR Methods}
\subsection{Gradient based guidance in T2I models}\label{sec:t2i_gradient}
\begin{figure}[htb]
\centering
\scriptsize
\resizebox{\linewidth}{!}{
\begin{tabular}{c}

\rotatebox{90}{~~\small \ti$-\Pi$GDM}{\includegraphics[width=0.235\linewidth]{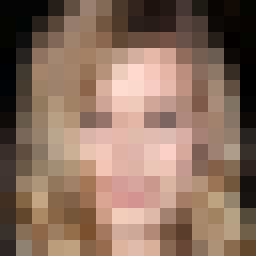}}
   \includegraphics[width=0.235\linewidth]{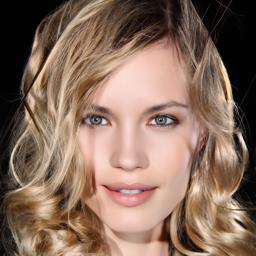}
\includegraphics[width=0.235\linewidth]{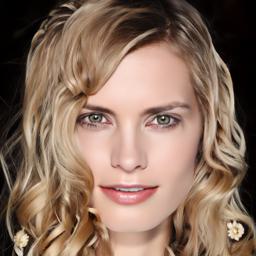}
\includegraphics[width=0.235\linewidth]{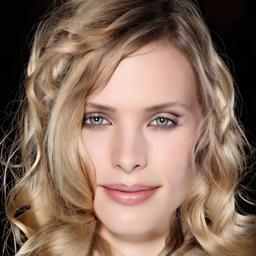}
\vspace{0.15em}\\
LR \qquad\qquad\quad 39.93dB\qquad\qquad\quad 41.72dB\qquad\qquad\quad 41.78dB\\
\rotatebox{90}{~~\small \ti$-\Pi$GDM}{\includegraphics[width=0.235\linewidth]{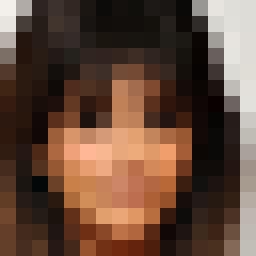}}
   \includegraphics[width=0.235\linewidth]{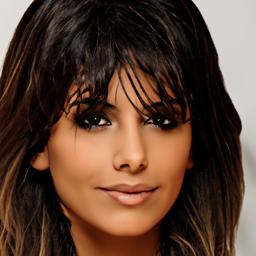}
\includegraphics[width=0.235\linewidth]{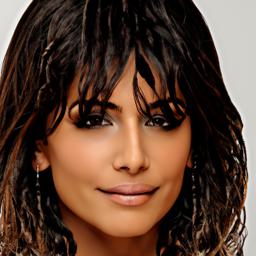}
\includegraphics[width=0.235\linewidth]{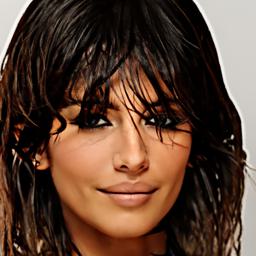}
\vspace{0.15em}\\
LR \qquad\qquad\quad 44.43dB\qquad\qquad\quad 43.28dB\qquad\qquad\quad 43.19dB\\
\hline
\vspace{0.15em}\\
\rotatebox{90}{~~~\small \ti$-$DPS}{\includegraphics[width=0.235\linewidth]{supplementary/pgdmx16/2_masked.jpg}}
   \includegraphics[width=0.235\linewidth]{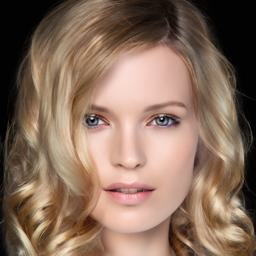}
\includegraphics[width=0.235\linewidth]{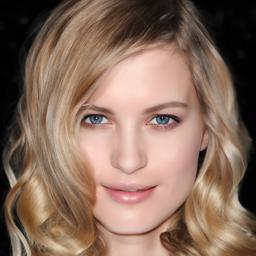}
\includegraphics[width=0.235\linewidth]{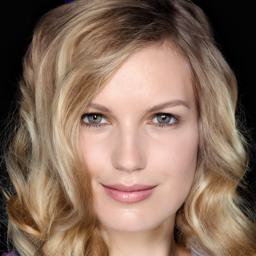}
\vspace{0.15em}\\
LR \qquad\qquad\quad 38.44dB\qquad\qquad\quad 38.94dB\qquad\qquad\quad 38.97dB\\
\rotatebox{90}{~~~\small \ti$-$DPS}{\includegraphics[width=0.235\linewidth]{supplementary/pgdmx16/5_masked.jpg}}
   \includegraphics[width=0.235\linewidth]{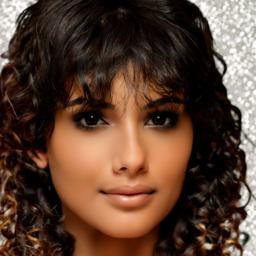}
\includegraphics[width=0.235\linewidth]{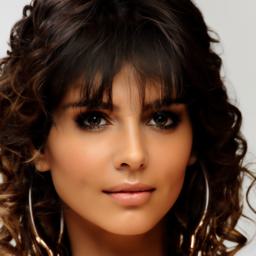}
\includegraphics[width=0.235\linewidth]{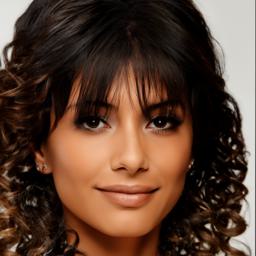}
\vspace{0.15em}\\
LR \qquad\qquad\quad 39.00dB\qquad\qquad\quad 39.02dB\qquad\qquad\quad 38.98dB\\
\end{tabular}}
\caption{Results of multiple runs of Imagen-$\Pi$GDM (top) and Imagen-DPS (bottom) with 1000 steps including classifier-free-guidance and corresponding LR PSNR values. Text prompt `Photograph of a face of a woman with curly hair'.\label{fig:pgdm_supp}}
\end{figure}

\begin{figure}[htb]
\centering
\scriptsize
\hspace{0.35\linewidth}\small{\bf Imagen$-\Pi$GDM}\hspace{0.08\linewidth} \small{\bf Imagen$-$DPS}\qquad\\
\includegraphics[width=0.325\linewidth]{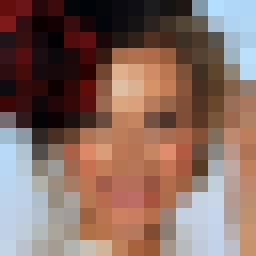}
   \includegraphics[width=0.325\linewidth]{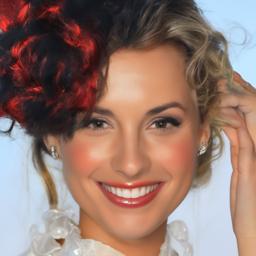}
\includegraphics[width=0.325\linewidth]{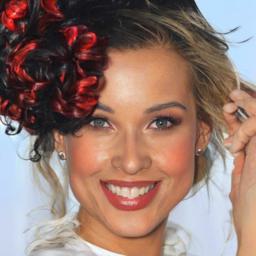}\\
LR \hspace{0.2\linewidth} 50.59dB\hspace{0.2\linewidth} 39.78dB\qquad\qquad\\
\includegraphics[width=0.325\linewidth]{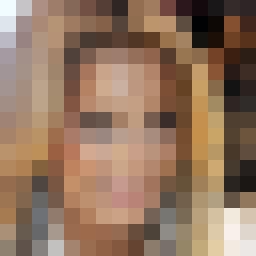}
   \includegraphics[width=0.325\linewidth]{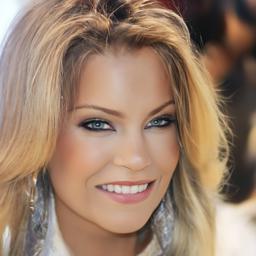}
\includegraphics[width=0.325\linewidth]{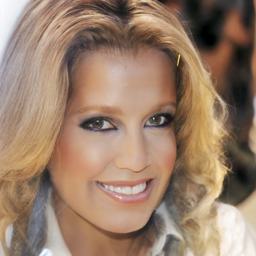}
LR \hspace{0.2\linewidth} 49.65dB\hspace{0.2\linewidth} 40.37dB\qquad\qquad\\
\includegraphics[width=0.325\linewidth]{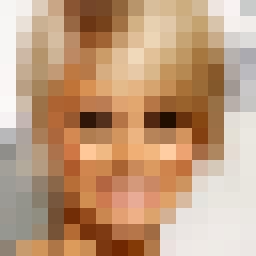}
   \includegraphics[width=0.325\linewidth]{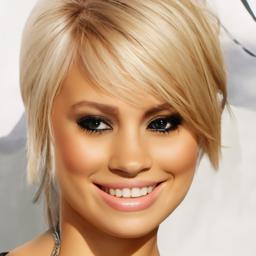}
\includegraphics[width=0.325\linewidth]{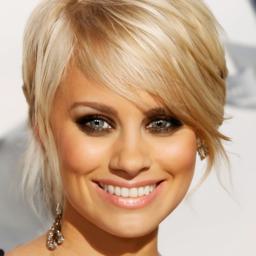}
LR \hspace{0.2\linewidth} 53.12dB\hspace{0.2\linewidth} 41.44dB\qquad\qquad\\
\includegraphics[width=0.325\linewidth]{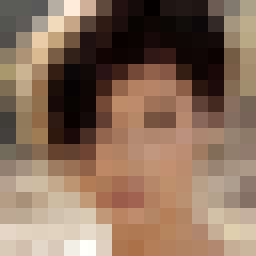}
   \includegraphics[width=0.325\linewidth]{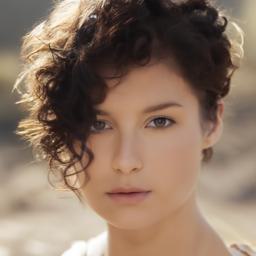}
\includegraphics[width=0.325\linewidth]{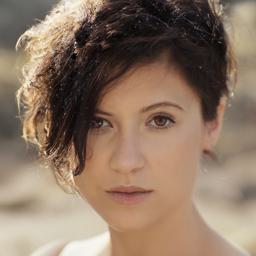}
LR \hspace{0.2\linewidth} 51.59dB\hspace{0.2\linewidth} 40.65dB\qquad\qquad\\
\includegraphics[width=0.325\linewidth]{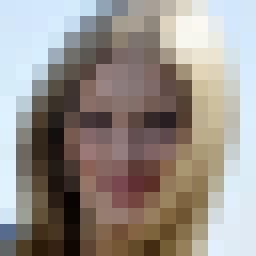}
   \includegraphics[width=0.325\linewidth]{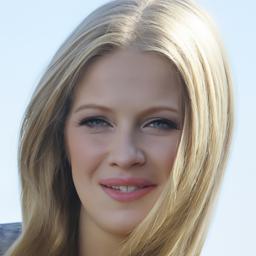}
\includegraphics[width=0.325\linewidth]{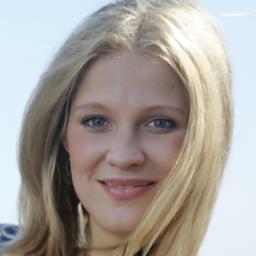}
LR \hspace{0.2\linewidth} 52.12dB\hspace{0.2\linewidth} 41.68dB\qquad\qquad\\
\caption{Example reconstructions of Imagen-$\Pi$GDM  and Imagen-DPS without classifier-free-guidance, and corresponding LR PSNR values. Text prompt `Photograph of the face of a woman with curly hair'\label{fig:pgdm_nocfg}.}
\end{figure}

We observed in \cref{tab:pigdm} of our paper that gradient-based measurement guidance could be in trade-off with the text
adherence. We found that increasing the number of diffusion steps to 1000 still did not improve LR consistency to the desired level ($>45$ dB) when we include classifier-free guidance for $\times 16$ super-resolution of faces.  We also performed experiments with Imagen+DPS, on a subset of 25 face images, for the text prompt 'a photograph of a woman with curly hair', the results are provided in \cref{tab:dps}.  While Imagen-$\Pi$GDM achieved the desired LR consistency without classifier-free guidance, we were unable to reach the desired level of LR consistency with Imagen$-$DPS, even without classifier-free guidance. When 1000 diffusion steps are used, we obtain an LR PSNR of around 40.9 dB. When classifier-free guidance is included, text adherence improves, with a significant drop in LR  PSNR. 
\begin{table}[]
    \centering
\resizebox{0.49\textwidth}{!}{\begin{tabular}{c|c|c|c|c|c}
\toprule
&Step size& Steps&CFG&LR PSNR (dB)&CLIP score\\
\midrule
 \multirow{4}{*}{Faces$16\times$}
 &1.0&500&\ding{55}&39.71&0.2433\\ 
  &1.0&800&\ding{55}&39.83&0.2409\\   
  &1.0&1000&\ding{55}&40.94&0.2353\\ 
   &0.5&500&\ding{51}&28.46& 0.3130\\
   &0.5&1000&\ding{51}&36.51& 0.2798\\
   \bottomrule
\end{tabular}}
    \caption{Effect of CFG, step size, and number of steps on consistency and text adherence in $\ti-DPS$ .\label{tab:dps}}
\end{table}
\cref{fig:pgdm_supp} shows sample results of this experiment for both  Imagen$-\Pi$GDM  and Imagen$-$DPS. Imagen$-\Pi$GDM obtains a higher PSNR than Imagen$-$DPS. \cref{fig:pgdm,fig:pgdm_supp} indicate that these methods can also achieve consistency with text at the cost of lower LR PSNR. We did not find solutions satisfying both the constraints in 3 independent runs. 
\cref{fig:pgdm_nocfg} shows sample reconstructions of Imagen$-\Pi$GDM  and Imagen$-$DPS when there is no classifier-free guidance for the same text prompt. We find that Imagen$-\Pi$GDM achieves significantly higher LR PSNR. While both methods achieve photo-realistic reconstructions, they fail to adhere to text prompts without classifier-free guidance.  

\begin{figure}[htb]
\centering
\scriptsize
\resizebox{\linewidth}{!}{
\begin{tabular}{c}
\includegraphics[width=0.235\linewidth]{images/sr_embd_avg/16xLR00032.jpg}
   \rotatebox{90}{~~~unCLIP  DDPM}{\includegraphics[width=0.235\linewidth]{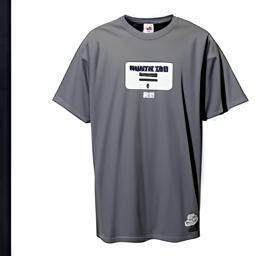}}
\includegraphics[width=0.235\linewidth]{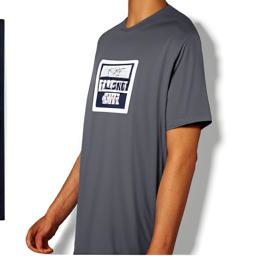}
\includegraphics[width=0.235\linewidth]{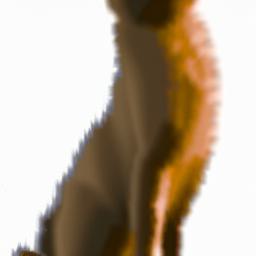}
\vspace{0.15em}\\
LR \qquad\qquad start 0 null-text\quad  start 10 null-text\qquad\quad pinv embed\\
\includegraphics[width=0.235\linewidth]{images/sr_embd_avg/16xLR00032.jpg}
   \rotatebox{90}{~~~unCLIP  DDNM}{\includegraphics[width=0.235\linewidth]{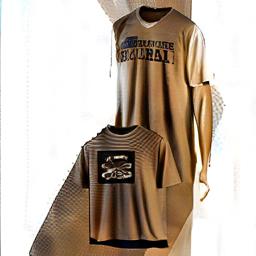}}
\includegraphics[width=0.235\linewidth]{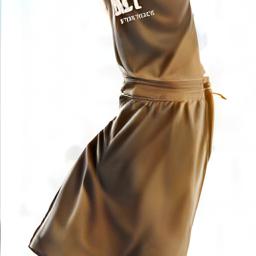}
\includegraphics[width=0.235\linewidth]{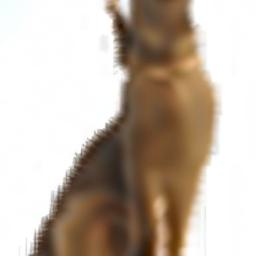}
\vspace{0.15em}\\
\qquad LR \qquad\qquad no embed avg\quad  embed avg $\lambda= 0.4$\quad embed avg $\lambda=1.0$\\
\end{tabular}}
\caption{Effect of image embedding in unCLIP-DDNM.\label{fig:unclip_im_embed}}
\end{figure}

\begin{figure}[t]
\centering
\scriptsize
 \hspace{-1.2em}{\rotatebox{90}{No averaging trick}} \includegraphics[width=0.33\linewidth]{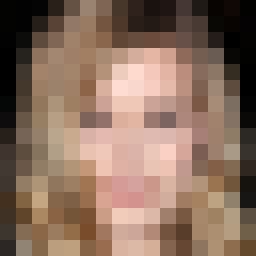}\hspace{-1.5pt}
\includegraphics[width=0.33\linewidth]{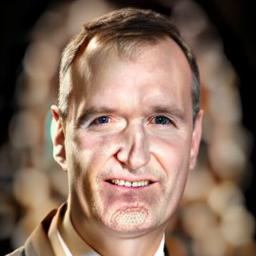}\hspace{-1.5pt}
\includegraphics[width=0.33\linewidth]{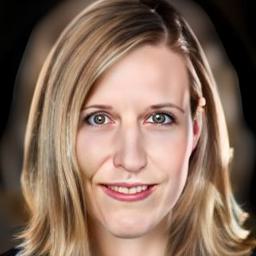}\vspace{0.15em}\\
\hspace{-1.2em}{\rotatebox{90}{with averaging trick}} \includegraphics[width=0.33\linewidth]{images/supple_embed/16xLR00002.jpg}\hspace{-1.5pt}
\includegraphics[width=0.33\linewidth]{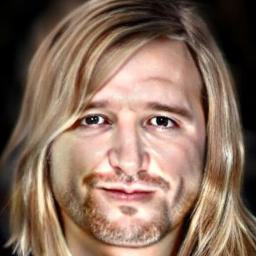}\hspace{-1.5pt}
\includegraphics[width=0.33\linewidth]{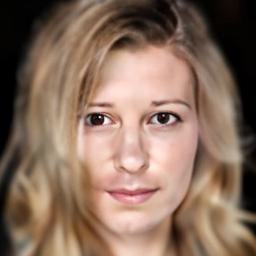}\vspace{0.15em}\\
\hspace{-1.2em}{\rotatebox{90}{No averaging trick}} \includegraphics[width=0.33\linewidth]{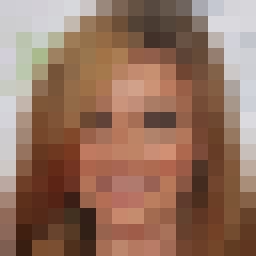}\hspace{-1.5pt}
\includegraphics[width=0.33\linewidth]{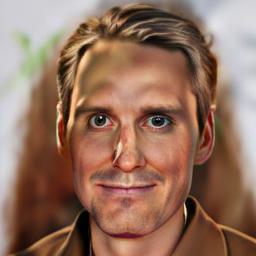}\hspace{-1.5pt}
\includegraphics[width=0.33\linewidth]{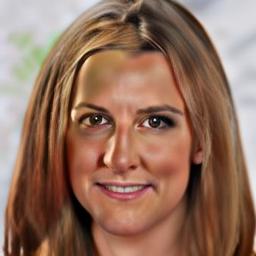}\vspace{0.15em}\\
\hspace{-1.2em}{\rotatebox{90}{with averaging trick}} 
\includegraphics[width=0.33\linewidth]{images/supple_embed/16xLR00046.jpg}\hspace{-1.5pt}
\includegraphics[width=0.33\linewidth]{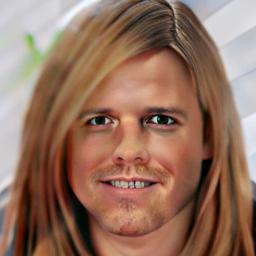}\hspace{-1.5pt}
\includegraphics[width=0.33\linewidth]{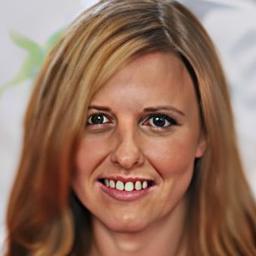}\vspace{0.15em}\\
 `A photograph of  a face of\{*\}' for $* \in$ \{a man, a woman\}
\caption{Embeddings averaging trick improves photorealism in difficult examples $\times16$ face super-resolution.\label{fig:faces_16xsr_avgtrick}}
\end{figure}

\paragraph{Influence of image embedding in unCLIP-DDNM} In our paper, we introduced an `embeddings averaging trick' and showed that this can reduce the mismatch between image embedding and the observation. We now illustrate the impact of image embedding on unCLIP-DDNM solutions with an example in \cref{fig:unclip_im_embed}. The first row shows the LR image, and the images generated by unCLIP through the usual DDPM sampling process with no text prompt, i.e. decoder is conditioned only on image embeddings generated by the prior with null-text. We fix the random seed throughout this experiment. This embedding produces an image of a T-shirt. Even starting the reverse diffusion process at an earlier step initializing with a pseudo-inverse solution retains the same concept. In the second row, we observe that using the prior image embedding as it is leads to a solution where the pseudoinverse is super-imposed on an image of T-shirts. On the other hand, embeddings averaging results align the concept provided by prior embedding with the pseudoinverse solution. The benefit of embeddings averaging for text-conditioned super-resolution is further illustrated in \cref{fig:faces_16xsr_avgtrick}. Even for text prompts that are seemingly incoherent with the low-resolution measurement such as `a photograph of a man', embeddings averaging can reduce misalignments and weird backgrounds and improve photorealism.
\subsection{Experiments with Stable Diffusion}
\paragraph{Using Text inputs in PSLD~\cite{rout2023solving}}
Recently \cite{rout2023solving} extends posterior sampling for inverse problems \cite{chung2022diffusion} to latent diffusion models such as Stable diffusion \cite{rombach2022high}. While this method also addresses super-resolution, they consider super-resolution with small SR factors $2\times,~3\times$, and $4\times$, and do not provide text prompts. Further, as Stablediffusion produces images at higher resolution $512\times512$, \cite{rout2023solving} apply the forward operator after upsampling inputs to
this resolution, run posterior sampling, and then downsample images to
resolution  $256\times256$. Therefore, even $4\times$ SR has a higher resolution input.  We instead experimented with a much lower input resolution of $32\times32$, which we interpolate to $64\times64$ and provide as input to PSLD for $8\times$ SR task. While the original setting of $4\times$ SR with $128\times128$ resolution provided satisfactory outputs,  we could not find a satisfactory hyper-parameter setting which results in a sharp output image with very low-resolution inputs.
\paragraph{Incorporating Null-space Consistency}
We also studied the applicability of Stable diffusion \cite{rombach2022high} for text-guided image super-resolution using DDNM. In the case of Stable Diffusion,  the diffusion process happens in the latent space of a variational auto-encoder, and it is not straightforward to adapt  DDNM to this model. 
\begin{figure}
 \centering
 \scriptsize 
  \includegraphics[width=0.33\linewidth]{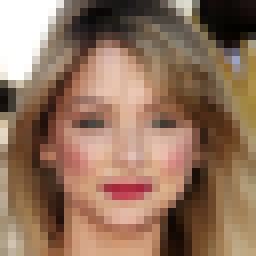}
  \hspace{-2pt}\includegraphics[width=0.33\linewidth]{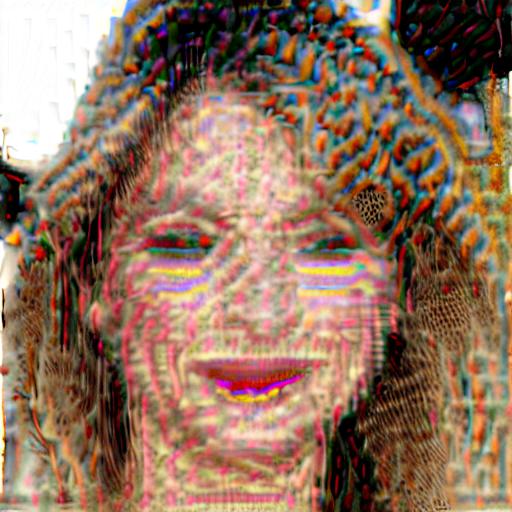}
  \hspace{-2pt}\includegraphics[width=0.33\linewidth]{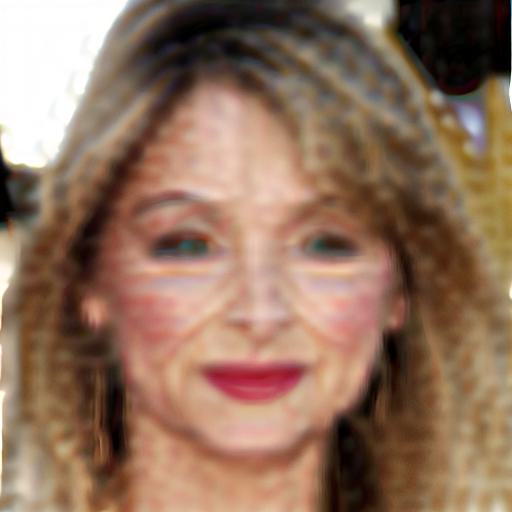}\\
  \scriptsize{\phantom{\Large AAA}LR\phantom{\Large AAAA} 40 DDNM steps \phantom{\Large AA}15 DDNM steps}\\
  \includegraphics[width=0.33\linewidth]{supplementary/supple_StableDiff/8xLR00020.jpg}
  \hspace{-2pt}\includegraphics[width=0.33\linewidth]{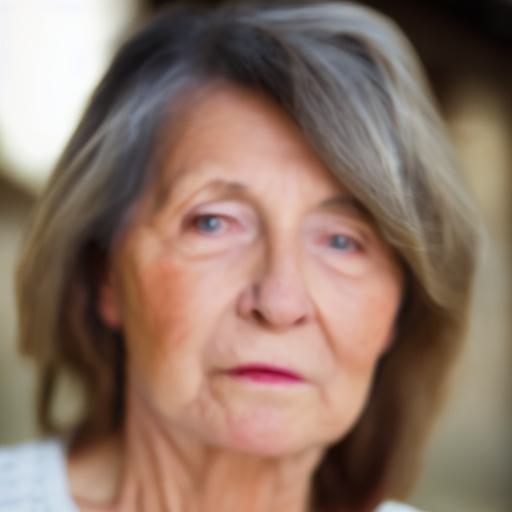}
  \hspace{-2pt}\includegraphics[width=0.33\linewidth]{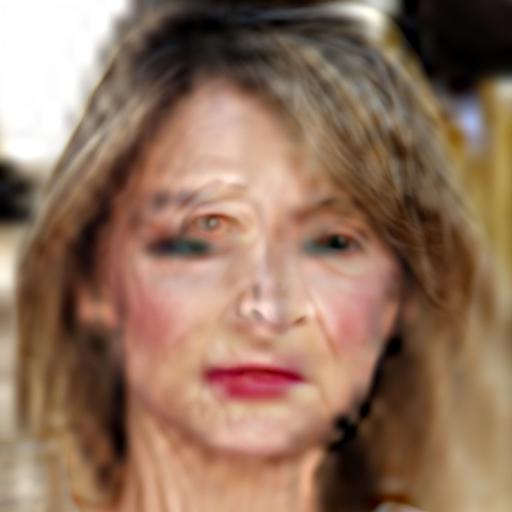}\\
  \scriptsize{\phantom{\Large AAA}LR\phantom{\Large AAAAA} SD output \phantom{\Large AAAl} After projection}
  \caption{(Top) Stable Diffusion with DDNM using VAE decoder \& encoder. (Bottom) Stable Diffusion results without DDNM in reverse diffusion. Results for the text prompt `a photograph of a face of an elderly woman' for $\times8$ SR}\label{fig:sd_ddnm}
  \end{figure}
We attempted to enforce DDNM consistency in the text-conditioned Stable diffusion. At each step in the reverse diffusion process, we estimate the clean latent variable $z_0$, and decode it to image space and enforce data consistency. This data-consistent image is then encoded again to latent space to resume reverse diffusion. While this approach achieves data-consistency, we find that it is highly unstable due to the lossy nature of the variational autoencoder. As the number of inference steps increases, it results in unrealistic images with heavy artifacts. When the number of inference steps is low, the artifacts reduce, however the resulting images are blurry, see the top row of Fig.~\ref{fig:sd_ddnm}. 
On the other hand, we observed that using the interpolated low-resolution image as an initial estimate in the diffusion process can lead to a totally different image without any guidance or consistency enforcement in the intermediate steps.   While the output of the Stable Diffusion model in this case is well-aligned with the text prompt, it is not consistent with the measurement. As a result, the corresponding null-space contents are not aligned with the pseudo-inverse solution. An example is illustrated in the bottom row of Fig.~\ref{fig:sd_ddnm}, where the null space projection adds high-frequency details of the elderly woman onto the pseudo inverse reconstruction. 
In contrast, using the pixel domain \ti models allows us to perform DDNM for text-guided reverse diffusion in the (down-sampled) pixel space, which easily generates data-consistent images that are aligned with text.
\subsection{{Noisy SR and Real-world SR:}} 
\begin{figure}
 \resizebox{0.99\linewidth}{!}{ \begin{tabular}{c}
    \includegraphics[width=0.15\linewidth]{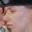}
    \includegraphics[width=0.15\linewidth]{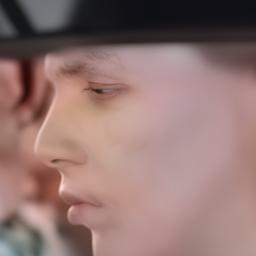}
    \includegraphics[width=0.15\linewidth]{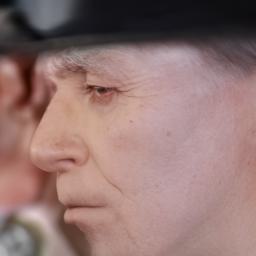}
    \includegraphics[width=0.15\linewidth]{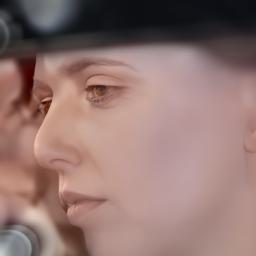}\vspace{-0.5em}\\
    \scriptsize{LR}\hspace{1cm}\scriptsize{`man'}\hspace{0.4cm}\scriptsize{`elderly man'}\hspace{0.3cm}\scriptsize{`woman'}\\
\end{tabular}}
\caption{Real world super-resolution on sample image from \cite{cheng2019low}.\label{fig:real}}
\end{figure}

We provide the results of $8\times$SR with Gaussian noise ($\sigma$=0.05) for the neutral prompt on CelebA HQ `a high-resolution photograph of a face'.  The methods $T2I$-DPS/$\Pi$GDM can directly handle noisy inputs. For $T2I$-DDNM we use DDNM\textsuperscript{+} following \cite{wang2022zero} and report LR PSNR, NIQE andCLIP score in \cref{tab:noisy}.
\begin{table}
\centering
    \resizebox{0.8\linewidth}{!}{ 
    \begin{tabular}{ccccc}
    \toprule
    Metric&DPS&\cite{saharia2022photorealistic}+DDNM\textsuperscript{+}&\cite{saharia2022photorealistic}+$\Pi$GDM&\cite{saharia2022photorealistic}+DPS\\
        \midrule
        LR PSNR&30.05 &31.40&31.52&31.33\\
        NIQE&5.92&7.09&6.07&5.57\\
        CLIP score&0.290&0.297&0.295&0.290\\
         \bottomrule
    \end{tabular}}
    \caption{Results of noisy super-resolution on CelebA HQ for $\times8$ SR.\label{tab:noisy}}
    \end{table}    
Zero-shot diffusion-based methods \cite{chung2022diffusion,wang2022zero,song2023pseudoinverseguided} require the knowledge of the degradation operator which is not available directly in real-world SR, but needs to be estimated. When the real LR images are well approximated by our downsampling operator we can still recover good quality solutions, see Fig.~\ref{fig:real} for results on real LR images from the QMUL tiny face dataset \cite{cheng2019low} using \cite{saharia2022photorealistic}+DDNM\textsuperscript{+}. 
\section{Experiment Settings}\label{sec:settings}
\paragraph{\ti+DDNM}
We investigate the use of Imagen \cite{saharia2022photorealistic} and unCLIP \cite{ramesh2022hierarchical}with DDNM sampling face super-resolution and open-domain image super-resolution.
 We use the open-source versions of Imagen Deep-Floyd IF~\cite{deepfloyd} and unCLIP  Karlo-unCLIP~\cite{kakaobrain2022karlo-v1-alpha}. We use these models for super-resolution without further fine-tuning. For Imagen+DDNM, we use 200 reverse diffusion steps at the resolution of $64\times64$, followed by 50 reverse diffusion steps at resolution $256\times256$.
 Instead of starting at random noise, we start reverse diffusion at timestamp $t_{stop}$ as 850, where noisy estimate at $t_{stop}$ is obtained by adding noise of suitable variance to the pseudo-inverse solution. We utilize classifier-free guidance with guidance scales of 7.0 and  4.0 respectively in the first and second stages of the Imagen Deep Floyd model. When compared to CFG using null-text, we found use of negative prompts \{ugly, disfigured, broken, caricature\} improves the quality of reconstruction. The guidance scale controls the strength of the attribute indicated by text and can be varied to obtain attributes of varying intensity. We found that not including classifier-free guidance can still produce realistic results, but the text adherence reduces significantly.

 For experiments with unCLIP+DDNM, we start reverse diffusion at timestamp $t_{stop}$ as 850 out of the total 1000 steps and use fewer steps between $[1, t_{stop}]$  in both the reverse diffusion process (80 for text conditioned decoding and 7 for super-resolution). The noisy estimate at $t_{stop}$ is obtained by adding noise of suitable variance to the pseudo-inverse solution. For obtaining the image embeddings we perform 25 reverse diffusion steps using the prior model.  We utilize an optional embeddings average trick when there is consistent misalignment between measurement and embeddings. For this, we utilize the CLIP image encoder used to train the unCLIP model.
 We utilize classifier-free guidance both for the prior and the unCLIP decoder. The super-resolution stage is conditioned only on the output of the first stage, and therefore, there is no classifier-free guidance in the SR module. We use classifier-free guidance of scale 4.0 in the prior. We find that a high value of classifier-free guidance scale for text conditioned decoder results in images which adhere well to text, yet they are often smoothed out, and appear unrealistic. We found lower values of classifier-free guidance between 1.0 and 3.0 leads to more photo-realistic results. Further, similar to Imagen+DDNM, we find better reconstructions when negative prompts are used for classifier-free guidance.

 While we provided hyperparameters used in experiments, these are not fixed and can be varied by the user to explore solutions. Oftentimes, a high number of reverse diffusion steps (200) is not required for Imagen+DDNM, and most solutions satisfying both text and low-resolution consistency can be obtained for fewer steps($\sim$100).
 All experiments with \ti+DDNM were run on a single GPU office machine, containing an NVIDIA RTX 3090 GPU.
\paragraph{CLIP guidance+DDNM}
We investigate the use of CLIP guidance along with DDNM for both face super-resolution and open-domain image super-resolution. For the task of face super-resolution, we use a diffusion model trained at a resolution of
$256\times256$ on CelebAMask-HQ dataset from available from SDEdit repository\footnote{\url{https://github.com/ermongroup/SDEdit}}. For the task of open domain image super-resolution, we use an unconditional imagenet model trained at a resolution of $256\times256$ from Guided Diffusion\footnote{\url{https://github.com/openai/guided-diffusion}}. We use the ViT-B/16 CLIP model for CLIP guidance. We incorporate DDNM null-space rectification into CLIP guided generation of \cite{yu2023freedom} using their publicly available  code\footnote{\url{https://github.com/vvictoryuki/FreeDoM}}. We retain their strategy for learning rate and time-travel strategy when using imagenet model for open domain super-resolution. We use 300 ddim steps with a learning rate hyperparameter of $0.05$.  Instead of $\ell_2$ distance used in \cite{yu2023freedom}, we utilize cosine distance measurement in the energy function for CLIP guidance.
All experiments with CLIP guidance were run on a single GPU office machine, containing an NVIDIA RTX 3090 GPU.
    \begin{figure*}[htb]
\centering
\scriptsize
  \includegraphics[width=0.19\linewidth]{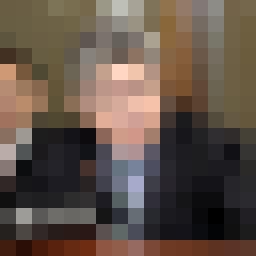}
\includegraphics[width=0.19\linewidth]{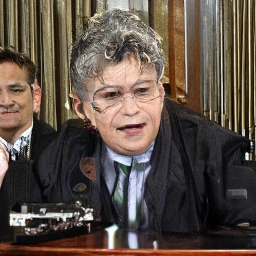}
\includegraphics[width=0.19\linewidth]{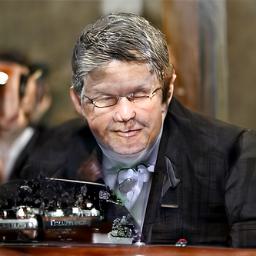}
\includegraphics[width=0.19\linewidth]{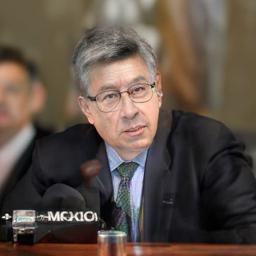}
\includegraphics[width=0.19\linewidth]{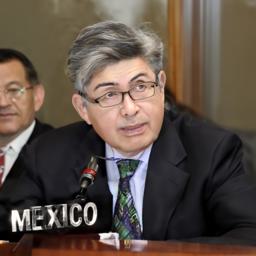}\\
A man with salt and pepper hair, a tie, and glasses is sitting behind a table with a sign that says Mexico in front of him.\\
\includegraphics[width=0.19\linewidth]{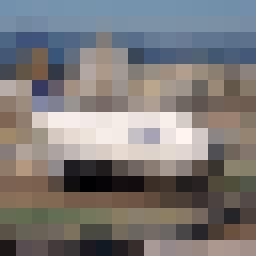}
\includegraphics[width=0.19\linewidth]{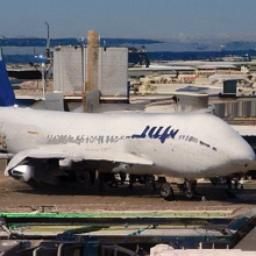}
\includegraphics[width=0.19\linewidth]{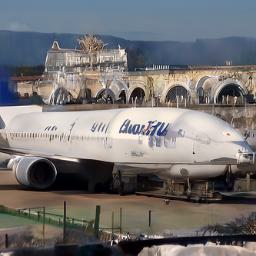}
\includegraphics[width=0.19\linewidth]{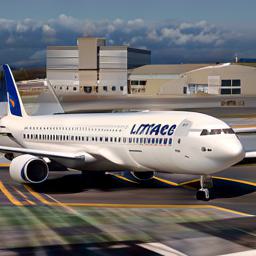}
\includegraphics[width=0.19\linewidth]{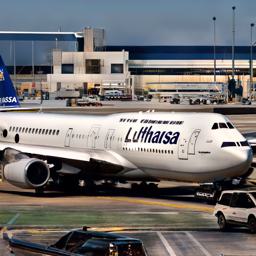}
\vspace{0.15em}\\
A large white Lufthansa plane is sitting on the runway.\\
\includegraphics[width=0.19\linewidth]{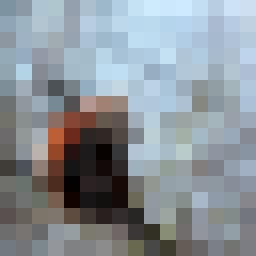}
\includegraphics[width=0.19\linewidth]{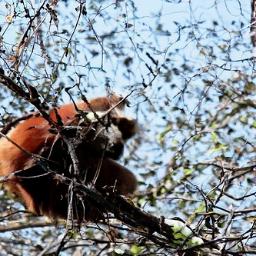}
\includegraphics[width=0.19\linewidth]{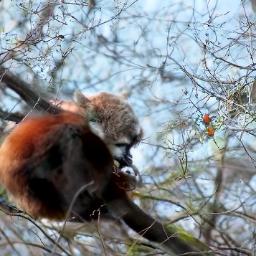}
\includegraphics[width=0.19\linewidth]{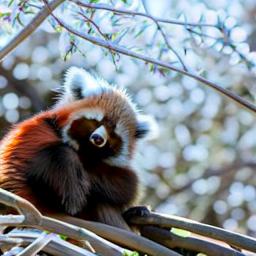}
\includegraphics[width=0.19\linewidth]{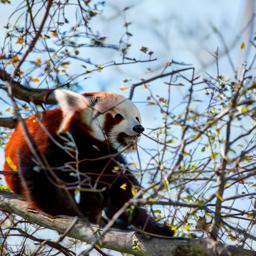}
\vspace{0.15em}\\
A red panda is sitting on a tree squeezing its eyes shut and sticking out its tongue.
\\
\includegraphics[width=0.19\linewidth]{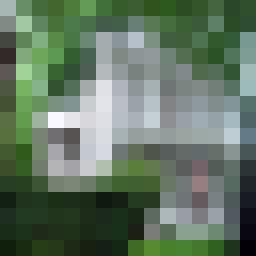}
\includegraphics[width=0.19\linewidth]{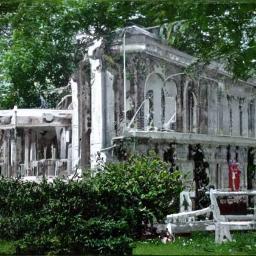}
\includegraphics[width=0.19\linewidth]{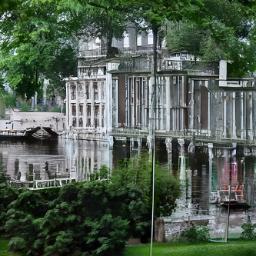}
\includegraphics[width=0.19\linewidth]{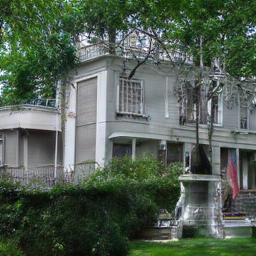}
\includegraphics[width=0.19\linewidth]{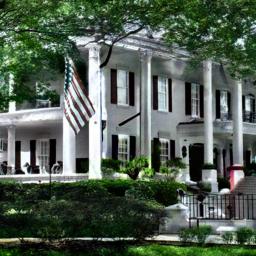}
\vspace{0.15em}\\
A stately grey and white house with an American flag sits among green trees.
\\
\includegraphics[width=0.19\linewidth]{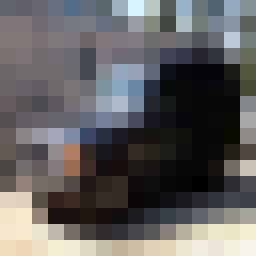}
\includegraphics[width=0.19\linewidth]{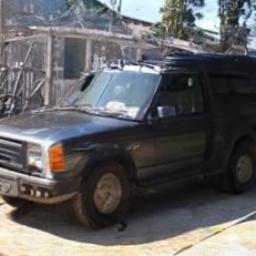}
\includegraphics[width=0.19\linewidth]{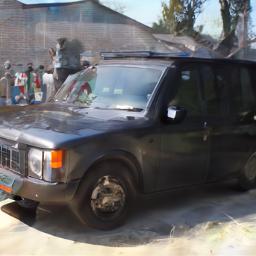}
\includegraphics[width=0.19\linewidth]{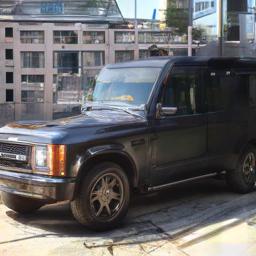}
\includegraphics[width=0.19\linewidth]{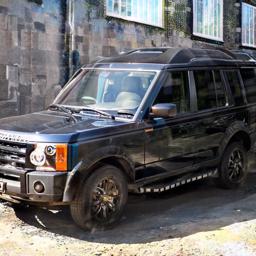}
\vspace{0.15em}\\
A black Land Rover is next to a building.
\\
\includegraphics[width=0.19\linewidth]{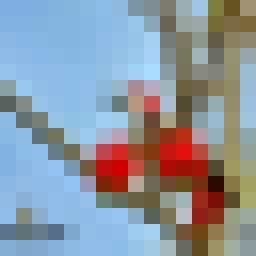}
\includegraphics[width=0.19\linewidth]{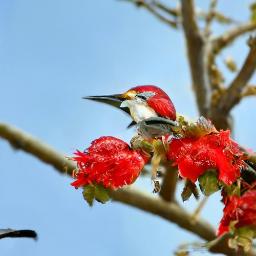}
\includegraphics[width=0.19\linewidth]{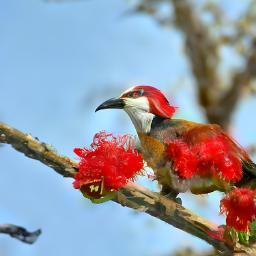}
\includegraphics[width=0.19\linewidth]{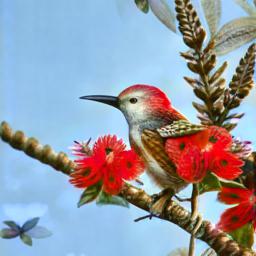}
\includegraphics[width=0.19\linewidth]{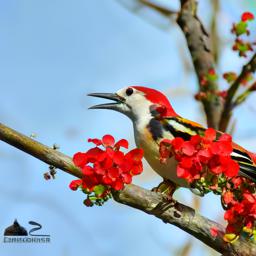}
\vspace{0.15em}\\
A gorgeous red headed woodpeker perched on a branch with red flowers.
\\
{\hspace*{0.08\linewidth}}\footnotesize{LR}{\hspace*{0.12\linewidth}}\footnotesize{DPS}\cite{chung2022diffusion}{\hspace*{0.08\linewidth}}\text{\footnotesize{DDNM+CLIP guidance} }{\hspace*{0.04\linewidth}}\text{\footnotesize{unCLIP\cite{ramesh2022hierarchical}+DDNM} }{\hspace*{0.06\linewidth}}\text{\footnotesize{Imagen\cite{saharia2022photorealistic}+DDNM} }
\caption{Visual comparison of $16\times$ SR on open domain images.\label{fig:supple_nocaps_sr}}
\end{figure*}
\begin{figure*}[htb]
\centering
\scriptsize
  \includegraphics[width=0.19\linewidth]{images/LR_imsx16/56_masked.jpg}
\includegraphics[width=0.19\linewidth]{images/DPS/56.jpg}
\includegraphics[width=0.19\linewidth]{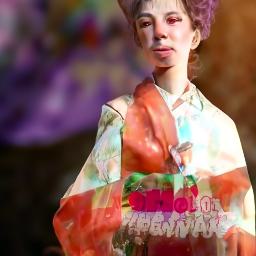}
\includegraphics[width=0.19\linewidth]{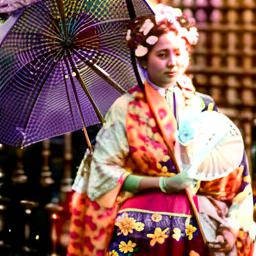}
\includegraphics[width=0.19\linewidth]{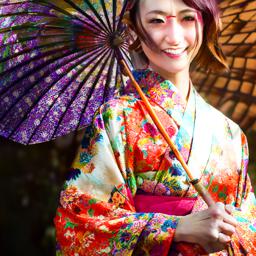}
\vspace{0.15em}\\
The woman is wearing a colorful kimono and carrying a floral print purple umbrella.\\
\includegraphics[width=0.19\linewidth]{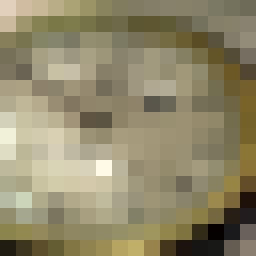}
\includegraphics[width=0.19\linewidth]{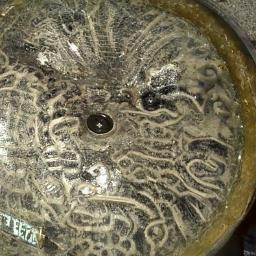}
\includegraphics[width=0.19\linewidth]{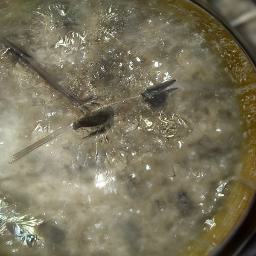}
\includegraphics[width=0.19\linewidth]{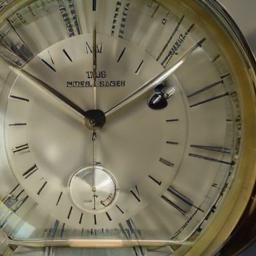}
\includegraphics[width=0.19\linewidth]{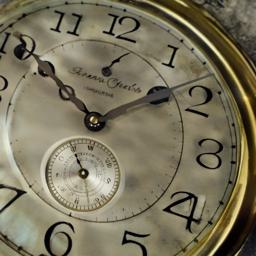}
\vspace{0.15em}\\
A clock with a gold bezel reading ten minutes to 2-o-clock.\\
\includegraphics[width=0.19\linewidth]{images/LR_imsx16/79_masked.jpg}
\includegraphics[width=0.19\linewidth]{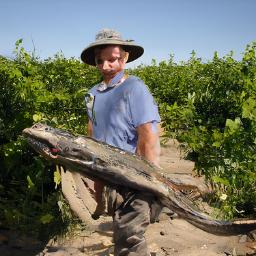}
\includegraphics[width=0.19\linewidth]{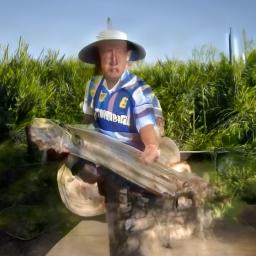}
\includegraphics[width=0.19\linewidth]{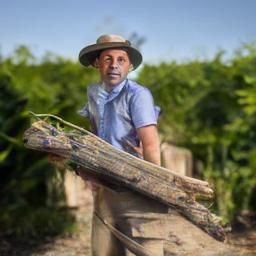}
\includegraphics[width=0.19\linewidth]{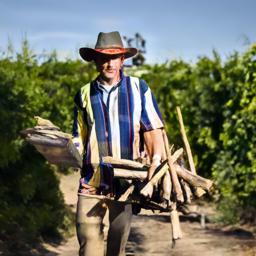}
\vspace{0.15em}\\
A man in a hat and blue striped shirt carrying wood\\
\includegraphics[width=0.19\linewidth]{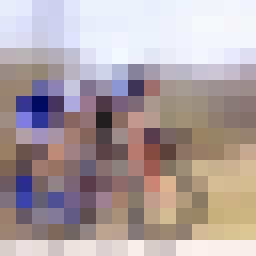}
\includegraphics[width=0.19\linewidth]{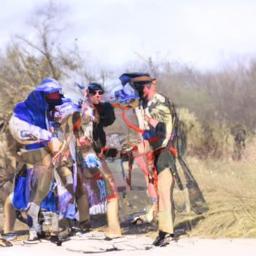}
\includegraphics[width=0.19\linewidth]{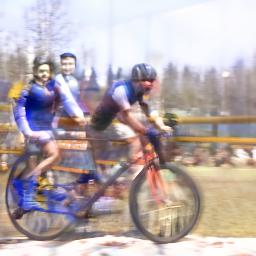}
\includegraphics[width=0.19\linewidth]{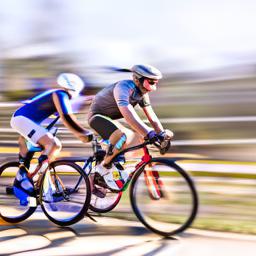}
\includegraphics[width=0.19\linewidth]{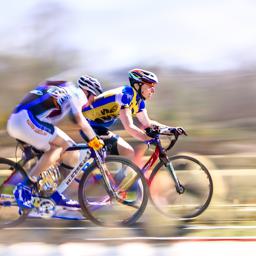}
\vspace{0.15em}\\
Two bicyclists are racing through a blurry background.
\\
\includegraphics[width=0.19\linewidth]{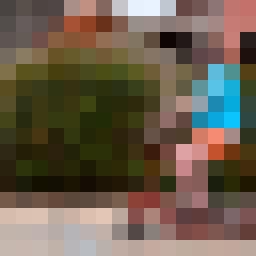}
\includegraphics[width=0.19\linewidth]{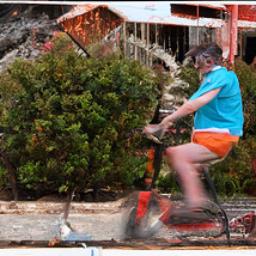}
\includegraphics[width=0.19\linewidth]{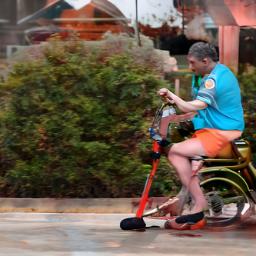}
\includegraphics[width=0.19\linewidth]{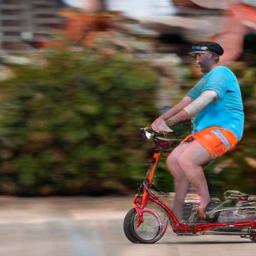}
\includegraphics[width=0.19\linewidth]{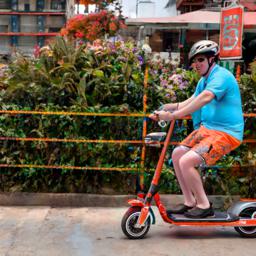}
\vspace{0.15em}\\
A man in a blue shirt and orange shorts ride a motorized scooter.
\\
\includegraphics[width=0.19\linewidth]{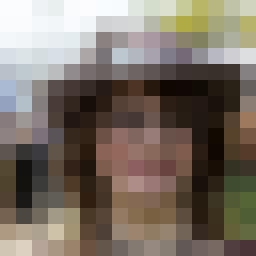}
\includegraphics[width=0.19\linewidth]{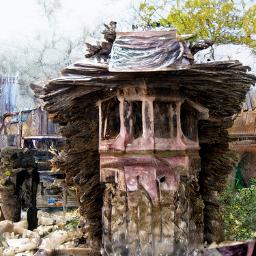}
\includegraphics[width=0.19\linewidth]{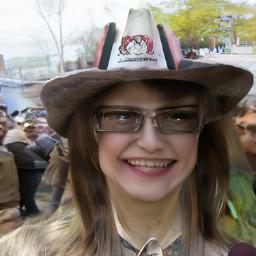}
\includegraphics[width=0.19\linewidth]{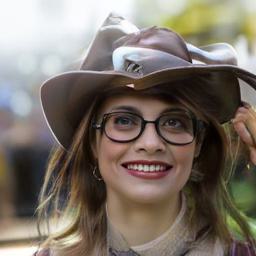}
\includegraphics[width=0.19\linewidth]{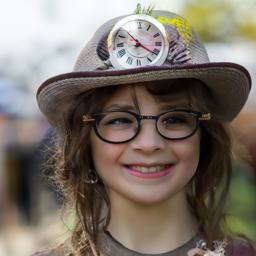}
\vspace{0.15em}\\
A smiling girl with glasses and a hat with a clock on it.
\\
{\hspace*{0.08\linewidth}}\footnotesize{LR}{\hspace*{0.12\linewidth}}\footnotesize{DPS}\cite{chung2022diffusion}{\hspace*{0.08\linewidth}}\text{\footnotesize{DDNM+CLIP guidance} }{\hspace*{0.04\linewidth}}\text{\footnotesize{unCLIP\cite{ramesh2022hierarchical}+DDNM} }{\hspace*{0.06\linewidth}}\text{\footnotesize{Imagen\cite{saharia2022photorealistic}+DDNM} }
\caption{Visual comparison of $16\times$ SR on open domain images.\label{fig:supple2_nocaps_sr}}
\end{figure*}
    \begin{figure*}
		\centering
    \small   
   \resizebox{\linewidth}{!}
{\begin{tabular}{cc}
  \multirow{5}{*}{
  \begin{tabular}{c}  
  \includegraphics[width=0.15\linewidth]{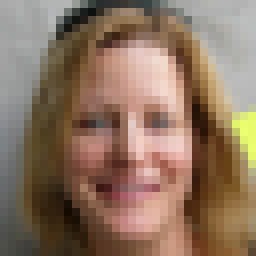}\\
  LR input\\
 \end{tabular}}&
      \includegraphics[width=0.21\linewidth]{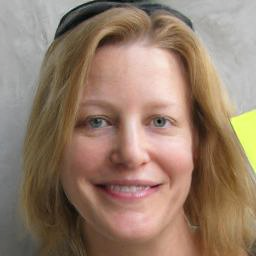} 
      \includegraphics[width=0.21\linewidth]{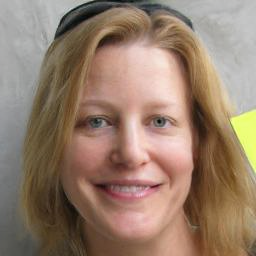}       
      \includegraphics[width=0.21\linewidth]{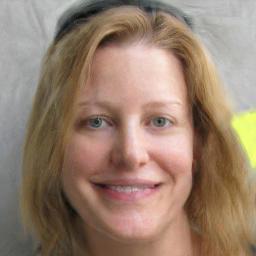}
      \includegraphics[width=0.21\linewidth]{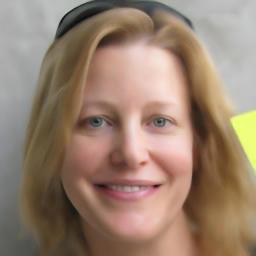}\\
      &
$\xleftarrow{\hspace*{0.15\linewidth}}\text{DPS \cite{chung2022diffusion}} \xrightarrow{\hspace*{0.15\linewidth}}\xleftarrow{\hspace*{0.15\linewidth}}\text{DDNM \cite{wang2022zero}} \xrightarrow{\hspace*{0.15\linewidth}}$\\
     & \hspace{-9pt} \rotatebox{90}{~~~~~~Imagen-DDNM}{    \includegraphics[width=0.21\linewidth]{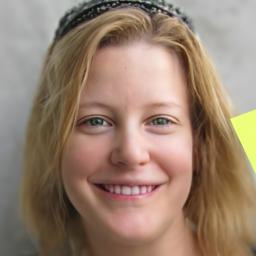}}     \includegraphics[width=0.21\linewidth]{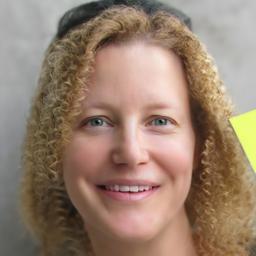} 
      \includegraphics[width=0.21\linewidth]{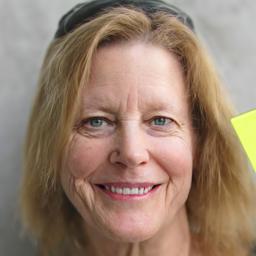}
      \includegraphics[width=0.21\linewidth]{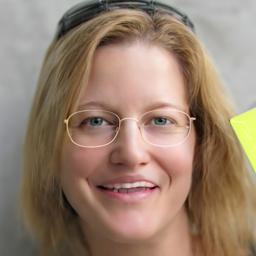}\\
         & \hspace{-9pt}\rotatebox{90}{~~~~~~unCLIP DDNM}{      \includegraphics[width=0.21\linewidth]{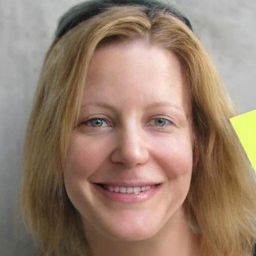}}     \includegraphics[width=0.21\linewidth]{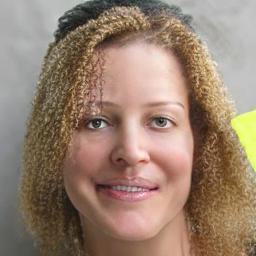} 
      \includegraphics[width=0.21\linewidth]{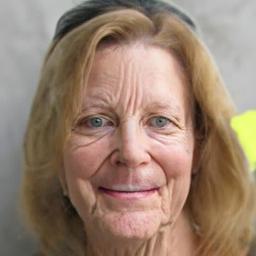}
      \includegraphics[width=0.21\linewidth]{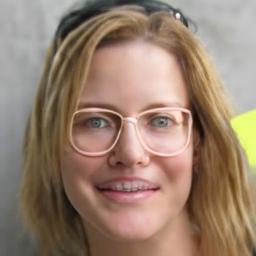}\\
      &\phantom{retsrzii}'a woman'\phantom{retsrzdtguji} 'a  woman with curly hair'\phantom{retsrz} 'an elderly woman'\phantom{retsrzd} 'a woman with glasses'\\
        \multirow{5}{*}{
  \begin{tabular}{c}  
  \includegraphics[width=0.15\linewidth]{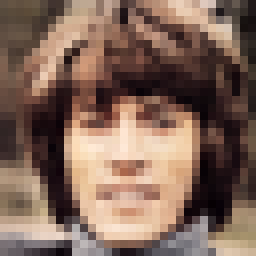}\\
  LR input\\
 \end{tabular}}&
      \includegraphics[width=0.21\linewidth]{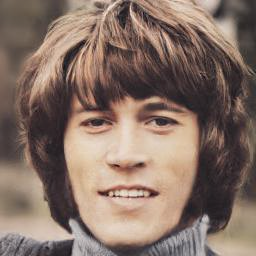} 
      \includegraphics[width=0.21\linewidth]{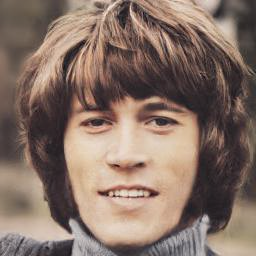}       
      \includegraphics[width=0.21\linewidth]{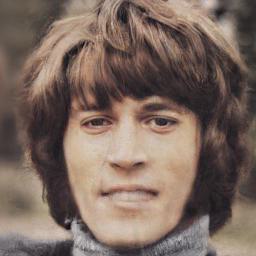}
      \includegraphics[width=0.21\linewidth]{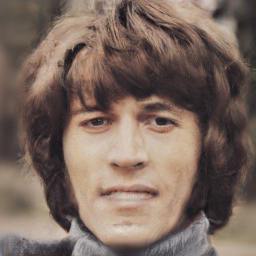}\\
      &
$\xleftarrow{\hspace*{0.15\linewidth}}\text{DPS \cite{chung2022diffusion}} \xrightarrow{\hspace*{0.15\linewidth}}\xleftarrow{\hspace*{0.15\linewidth}}\text{DDNM \cite{wang2022zero}} \xrightarrow{\hspace*{0.15\linewidth}}$\\
     & \hspace{-9pt}    \rotatebox{90}{~~~~~~Imagen-DDNM}{  \includegraphics[width=0.21\linewidth]{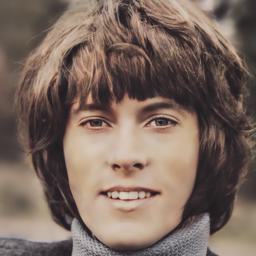} }      \includegraphics[width=0.21\linewidth]{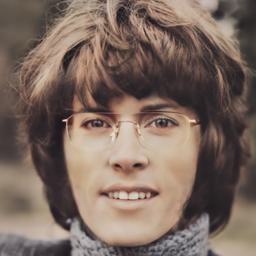} 
      \includegraphics[width=0.21\linewidth]{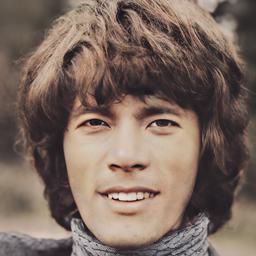}
      \includegraphics[width=0.21\linewidth]{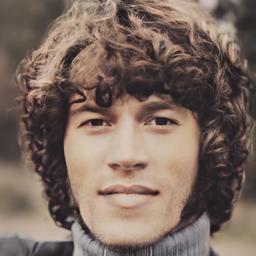}\\
        & \hspace{-9pt}  \rotatebox{90}{~~~~~~unCLIP DDNM}{    \includegraphics[width=0.21\linewidth]{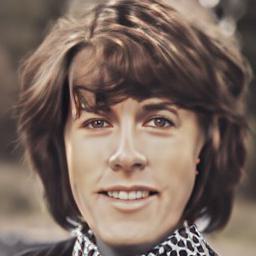} }      \includegraphics[width=0.21\linewidth]{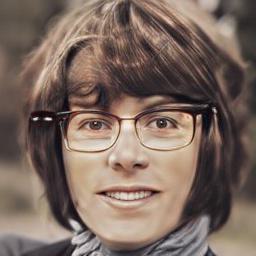} 
      \includegraphics[width=0.21\linewidth]{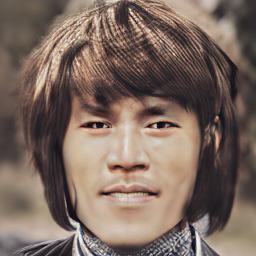}
      \includegraphics[width=0.21\linewidth]{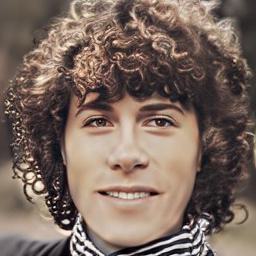}\\
      &\phantom{retsrzii}'a woman'\phantom{retsrzdtgi} 'a  woman with glasses'\phantom{retsrzdtuji} 'an Asian man'\phantom{retsrzd} 'a man with curly hair'\\
      \end{tabular}}    
    		\caption{Exploring solutions to $8\times$ face super-resolution }
		\label{fig:8xsr_supple}
\end{figure*}
\begin{figure*}
		\centering
    \small   
   \resizebox{\linewidth}{!}
{\begin{tabular}{cc}
  \multirow{5}{*}{
  \begin{tabular}{c}  
  \includegraphics[width=0.15\linewidth]{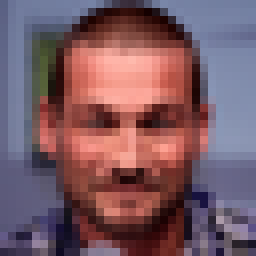}\\
  LR input\\
 \end{tabular}}&
      \includegraphics[width=0.21\linewidth]{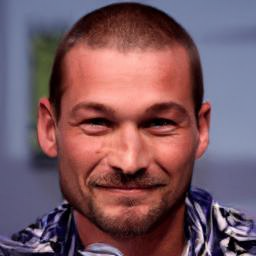} 
      \includegraphics[width=0.21\linewidth]{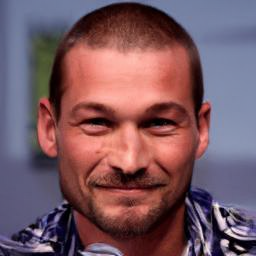}       
      \includegraphics[width=0.21\linewidth]{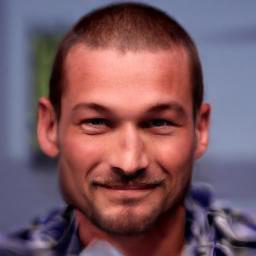}
      \includegraphics[width=0.21\linewidth]{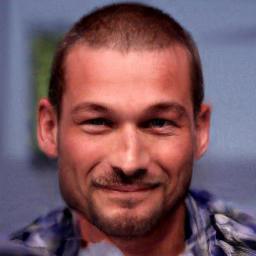}\\
      &
$\xleftarrow{\hspace*{0.15\linewidth}}\text{DPS \cite{chung2022diffusion}} \xrightarrow{\hspace*{0.15\linewidth}}\xleftarrow{\hspace*{0.15\linewidth}}\text{DDNM \cite{wang2022zero}} \xrightarrow{\hspace*{0.15\linewidth}}$\\
     & \hspace{-9pt} \rotatebox{90}{~~~~~~Imagen-DDNM}{    \includegraphics[width=0.21\linewidth]{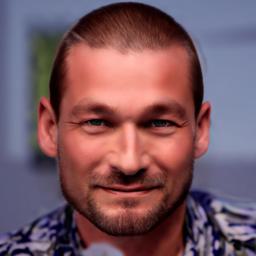}}     \includegraphics[width=0.21\linewidth]{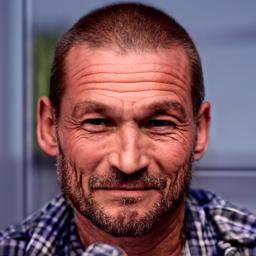} 
      \includegraphics[width=0.21\linewidth]{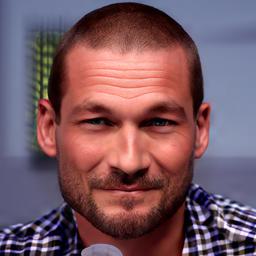}
      \includegraphics[width=0.21\linewidth]{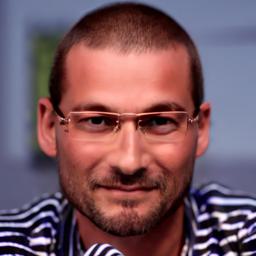}\\
         & \hspace{-9pt}\rotatebox{90}{~~~~~~unCLIP DDNM}{      \includegraphics[width=0.21\linewidth]{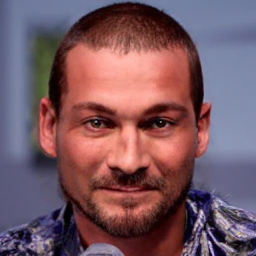}}     \includegraphics[width=0.21\linewidth]{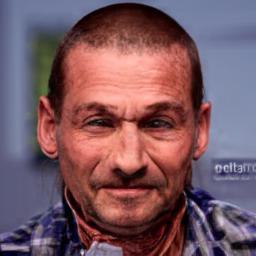} 
      \includegraphics[width=0.21\linewidth]{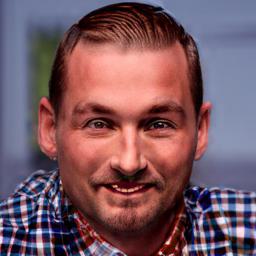}
      \includegraphics[width=0.21\linewidth]{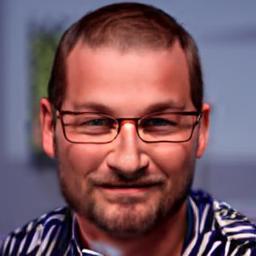}\\
      &\phantom{retsrzii}'a man'\phantom{retsrzdtgujiguji} 'an  elderly man'\phantom{retsrzdujit} 'man+checkered shirt '\phantom{retujis}'man+striped shirt+glasses'\\
      \multirow{5}{*}{
  \begin{tabular}{c}  
  \includegraphics[width=0.15\linewidth]{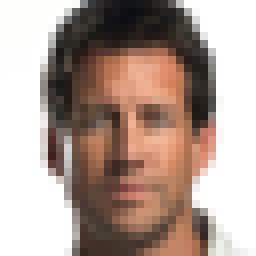}\\
  LR input\\
 \end{tabular}}&
      \includegraphics[width=0.21\linewidth]{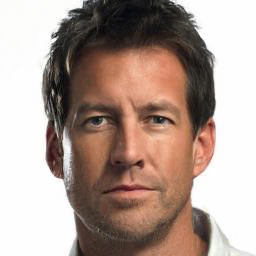} 
      \includegraphics[width=0.21\linewidth]{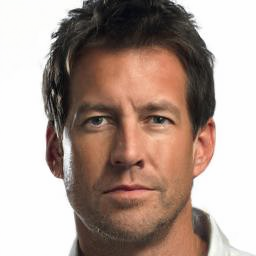}       
      \includegraphics[width=0.21\linewidth]{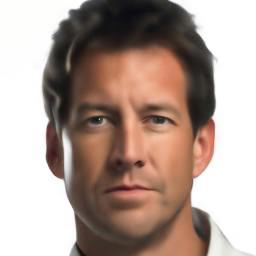}
      \includegraphics[width=0.21\linewidth]{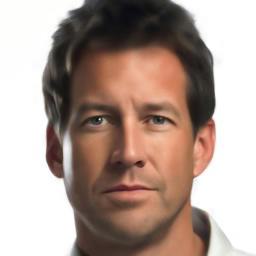}\\
      &
$\xleftarrow{\hspace*{0.15\linewidth}}\text{DPS \cite{chung2022diffusion}} \xrightarrow{\hspace*{0.15\linewidth}}\xleftarrow{\hspace*{0.15\linewidth}}\text{DDNM \cite{wang2022zero}} \xrightarrow{\hspace*{0.15\linewidth}}$\\
     & \hspace{-9pt} \rotatebox{90}{~~~~~~Imagen-DDNM}{    \includegraphics[width=0.21\linewidth]{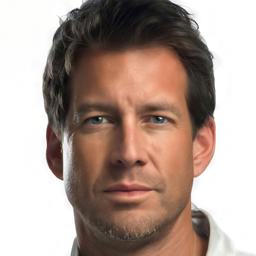}}     \includegraphics[width=0.21\linewidth]{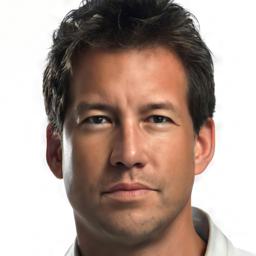} 
      \includegraphics[width=0.21\linewidth]{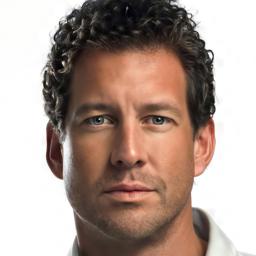}
      \includegraphics[width=0.21\linewidth]{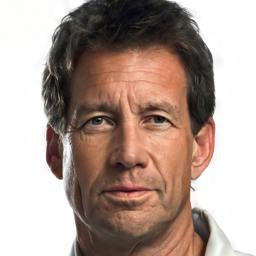}\\
         & \hspace{-9pt}\rotatebox{90}{~~~~~~unCLIP DDNM}{      \includegraphics[width=0.21\linewidth]{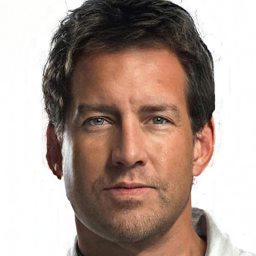}}     \includegraphics[width=0.21\linewidth]{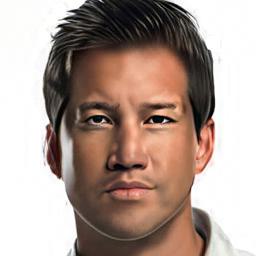} 
      \includegraphics[width=0.21\linewidth]{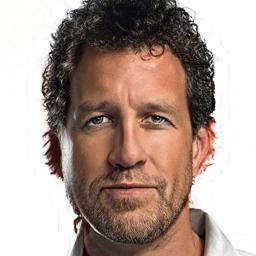}
      \includegraphics[width=0.21\linewidth]{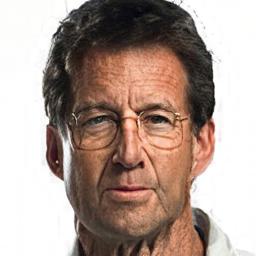}\\
      &\phantom{retsrzii}'a man'\phantom{retsrzdtgujiguji} 'an Asian man'\phantom{retsrzdujit} 'a man with curly hair '\phantom{retujis}'an elderly man'\\
      \end{tabular}}    
    		\caption{Exploring solutions to $8\times$ face super-resolution.}
		\label{fig:8xsr_supple_2}
\end{figure*}
    \begin{figure*}[htb]
\centering
\scriptsize
  \includegraphics[width=0.151\linewidth]{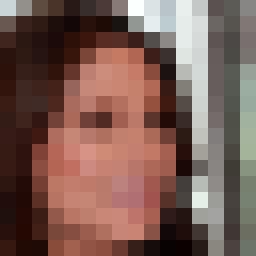}
\includegraphics[width=0.151\linewidth]{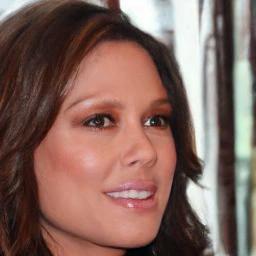}
\includegraphics[width=0.151\linewidth]{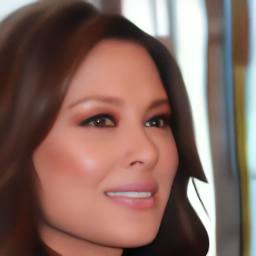}
\includegraphics[width=0.151\linewidth]{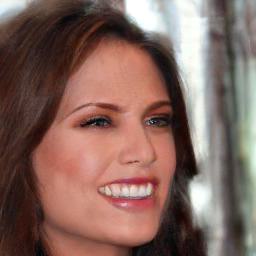}
\includegraphics[width=0.151\linewidth]{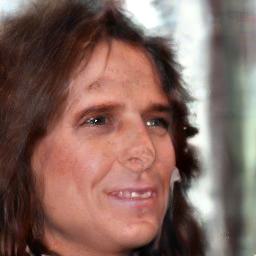}
\includegraphics[width=0.151\linewidth]{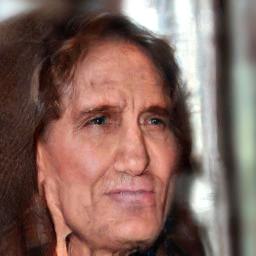}\\
\phantom{AAA}\footnotesize{LR}\phantom{AAAAAAAAAA}\footnotesize{DPS}\cite{chung2022diffusion}\phantom{AAAAAAAAAAA}DDNM\cite{wang2022zero}\phantom{AAA}$\xleftarrow{\hspace*{0.13\linewidth}}\text{CLIP guidance+DDNM} \xrightarrow{\hspace*{0.13\linewidth}}$\\
\includegraphics[width=0.151\linewidth]{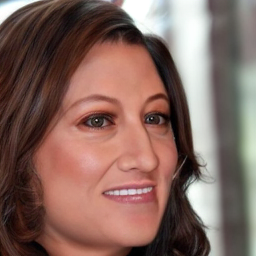}
\includegraphics[width=0.151\linewidth]{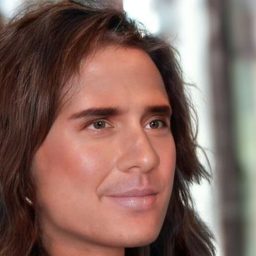}
\includegraphics[width=0.151\linewidth]{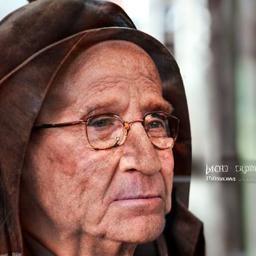}
\includegraphics[width=0.151\linewidth]{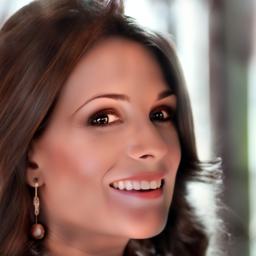}
\includegraphics[width=0.151\linewidth]{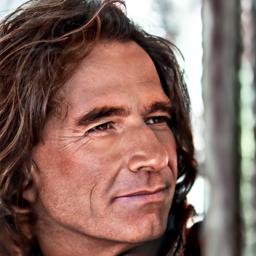}
\includegraphics[width=0.151\linewidth]{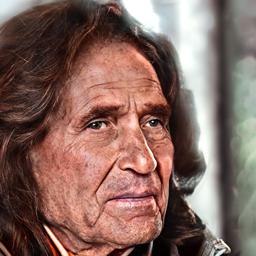}
 \\$\xleftarrow{\hspace*{0.17\linewidth}}\text{\cite{ramesh2022hierarchical}+DDNM} \xrightarrow{\hspace*{0.17\linewidth}}\xleftarrow{\hspace*{0.17\linewidth}}\text{\cite{saharia2022photorealistic}+DDNM} \xrightarrow{\hspace*{0.17\linewidth}}$\\
Text prompts: `Smiling woman', ` Man', `Elderly man'\\
  \includegraphics[width=0.151\linewidth]{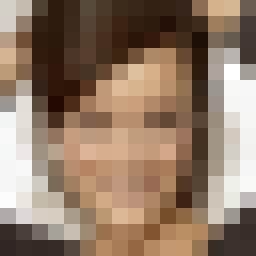}
\includegraphics[width=0.151\linewidth]{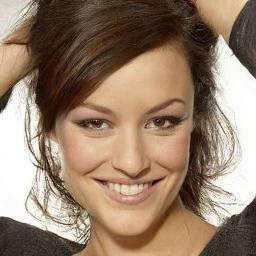}
\includegraphics[width=0.151\linewidth]{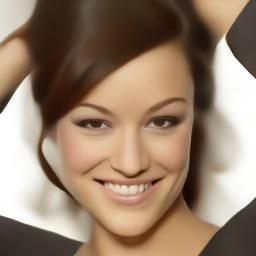}
\includegraphics[width=0.151\linewidth]{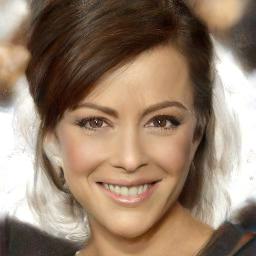}
\includegraphics[width=0.151\linewidth]{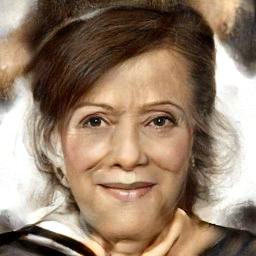}
\includegraphics[width=0.151\linewidth]{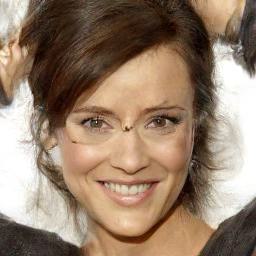}\\
\phantom{AAA}\footnotesize{LR}\phantom{AAAAAAAAAA}\footnotesize{DPS}\cite{chung2022diffusion}\phantom{AAAAAAAAAAA}DDNM\cite{wang2022zero}\phantom{AAA}$\xleftarrow{\hspace*{0.13\linewidth}}\text{CLIP guidance+DDNM} \xrightarrow{\hspace*{0.13\linewidth}}$\\
\includegraphics[width=0.151\linewidth]{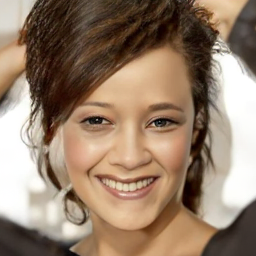}
\includegraphics[width=0.151\linewidth]{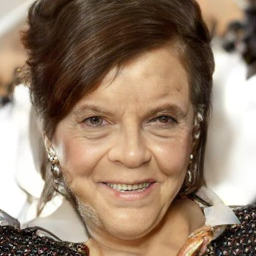}
\includegraphics[width=0.151\linewidth]{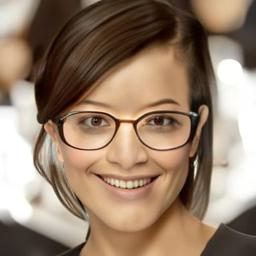}
\includegraphics[width=0.151\linewidth]{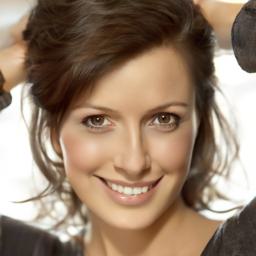}
\includegraphics[width=0.151\linewidth]{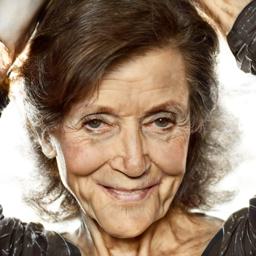}
\includegraphics[width=0.151\linewidth]{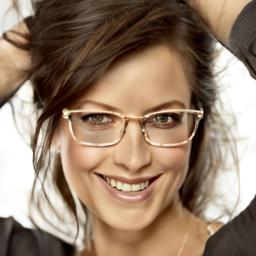}
 \\$\xleftarrow{\hspace*{0.17\linewidth}}\text{\cite{ramesh2022hierarchical}+DDNM} \xrightarrow{\hspace*{0.17\linewidth}}\xleftarrow{\hspace*{0.17\linewidth}}\text{\cite{saharia2022photorealistic}+DDNM} \xrightarrow{\hspace*{0.17\linewidth}}$\\
Text prompts: `Smiling woman', `Smiling elderly woman',`Smiling woman with glasses'\\

  \includegraphics[width=0.151\linewidth]{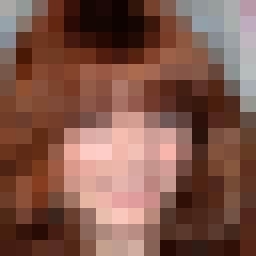}
\includegraphics[width=0.151\linewidth]{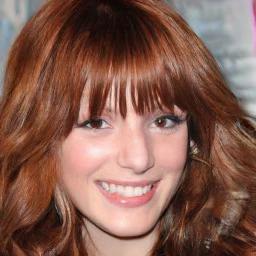}
\includegraphics[width=0.151\linewidth]{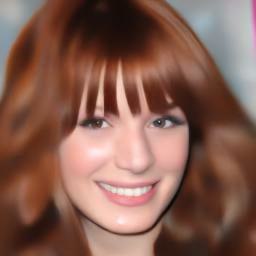}
\includegraphics[width=0.151\linewidth]{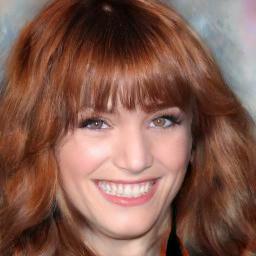}
\includegraphics[width=0.151\linewidth]{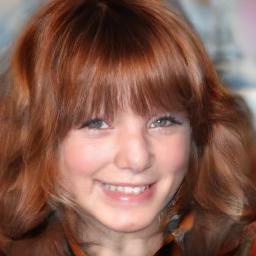}
\includegraphics[width=0.151\linewidth]{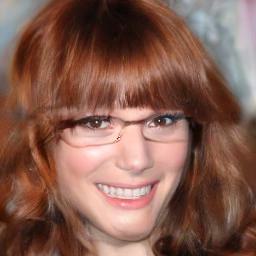}\\
\phantom{AAA}\footnotesize{LR}\phantom{AAAAAAAAAA}\footnotesize{DPS}\cite{chung2022diffusion}\phantom{AAAAAAAAAAA}DDNM\cite{wang2022zero}\phantom{AAA}$\xleftarrow{\hspace*{0.13\linewidth}}\text{CLIP guidance+DDNM} \xrightarrow{\hspace*{0.13\linewidth}}$\\
\includegraphics[width=0.151\linewidth]{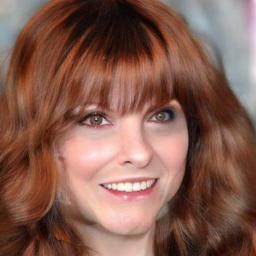}
\includegraphics[width=0.151\linewidth]{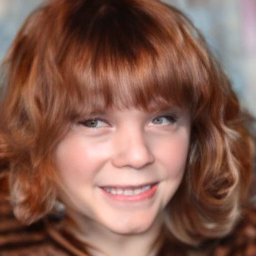}
\includegraphics[width=0.151\linewidth]{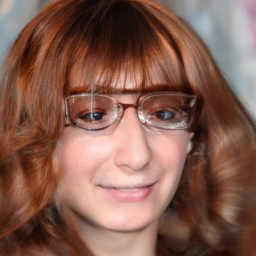}
\includegraphics[width=0.151\linewidth]{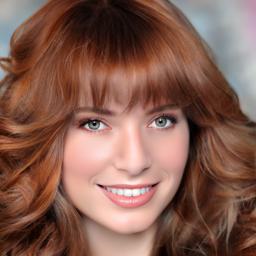}
\includegraphics[width=0.151\linewidth]{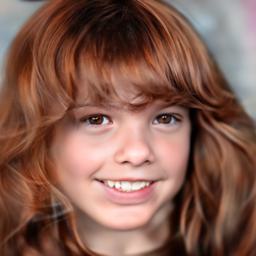}
\includegraphics[width=0.151\linewidth]{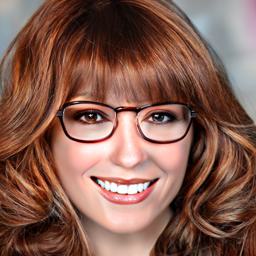}
 \\$\xleftarrow{\hspace*{0.17\linewidth}}\text{\cite{ramesh2022hierarchical}+DDNM} \xrightarrow{\hspace*{0.17\linewidth}}\xleftarrow{\hspace*{0.17\linewidth}}\text{\cite{saharia2022photorealistic}+DDNM} \xrightarrow{\hspace*{0.17\linewidth}}$\\
Text prompts: `Smiling woman', ` Smiling child', `Smiling woman with glasses'\\
\caption{Exploring solutions for $16\times$ SR of face images.\label{fig:supple16xfaces}}
\end{figure*}

\begin{figure*}[htb]
\centering
\scriptsize
  \includegraphics[width=0.151\linewidth]{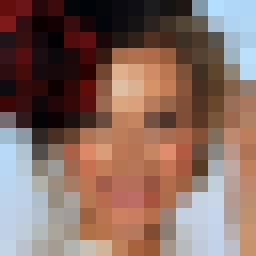}
\includegraphics[width=0.151\linewidth]{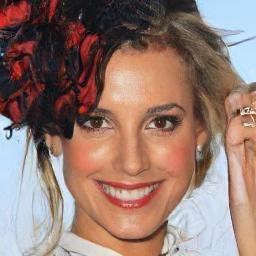}
\includegraphics[width=0.151\linewidth]{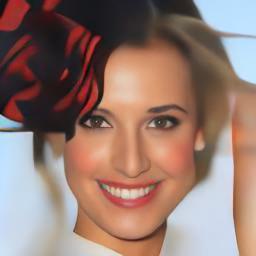}
\includegraphics[width=0.151\linewidth]{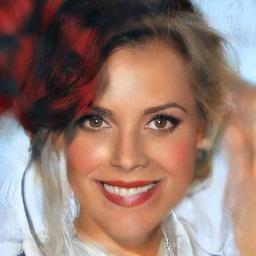}
\includegraphics[width=0.151\linewidth]{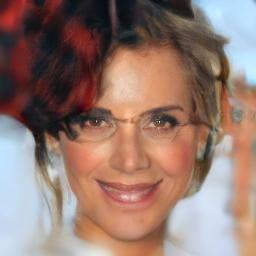}
\includegraphics[width=0.151\linewidth]{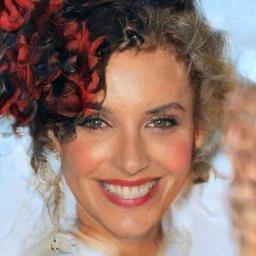}\\
\phantom{AAA}\footnotesize{LR}\phantom{AAAAAAAAAA}\footnotesize{DPS}\cite{chung2022diffusion}\phantom{AAAAAAAAAAA}DDNM\cite{wang2022zero}\phantom{AAA}$\xleftarrow{\hspace*{0.13\linewidth}}\text{CLIP guidance+DDNM} \xrightarrow{\hspace*{0.13\linewidth}}$\\
\includegraphics[width=0.151\linewidth]{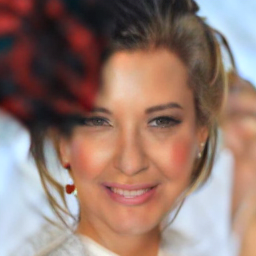}
\includegraphics[width=0.151\linewidth]{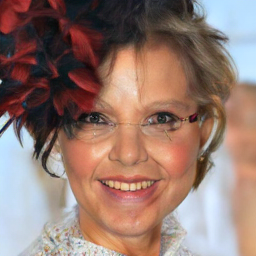}
\includegraphics[width=0.151\linewidth]{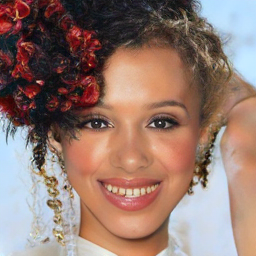}
\includegraphics[width=0.151\linewidth]{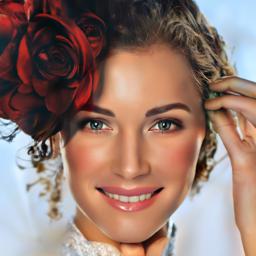}
\includegraphics[width=0.151\linewidth]{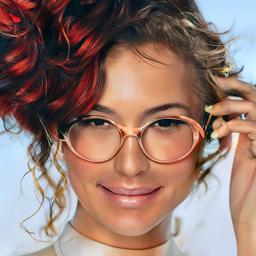}
\includegraphics[width=0.151\linewidth]{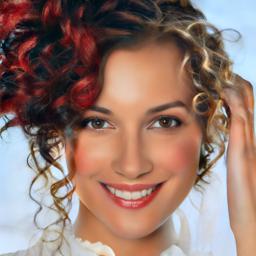}
 \\$\xleftarrow{\hspace*{0.17\linewidth}}\text{\cite{ramesh2022hierarchical}+DDNM} \xrightarrow{\hspace*{0.17\linewidth}}\xleftarrow{\hspace*{0.17\linewidth}}\text{\cite{saharia2022photorealistic}+DDNM} \xrightarrow{\hspace*{0.17\linewidth}}$\\
Text prompts: `Smiling woman', ` Woman with glasses', `Smiling woman with curly hair'\\
  \includegraphics[width=0.151\linewidth]{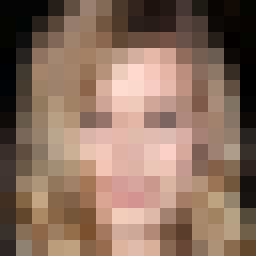}
\includegraphics[width=0.151\linewidth]{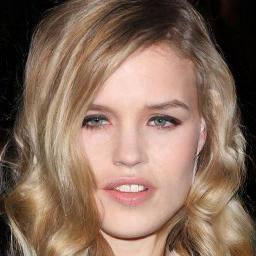}
\includegraphics[width=0.151\linewidth]{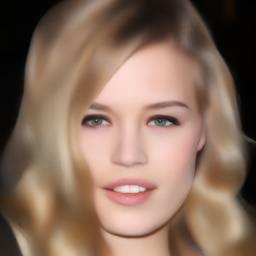}
\includegraphics[width=0.151\linewidth]{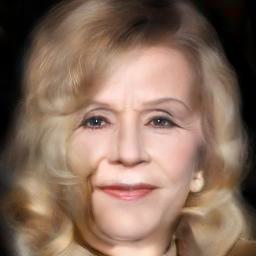}
\includegraphics[width=0.151\linewidth]{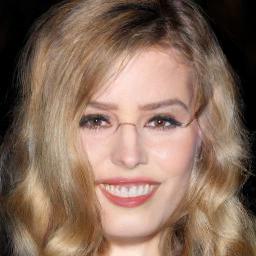}
\includegraphics[width=0.151\linewidth]{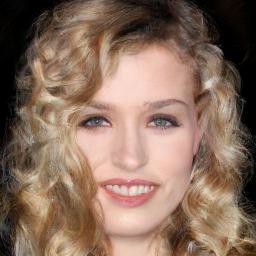}\\
\phantom{AAA}\footnotesize{LR}\phantom{AAAAAAAAAA}\footnotesize{DPS}\cite{chung2022diffusion}\phantom{AAAAAAAAAAA}DDNM\cite{wang2022zero}\phantom{AAA}$\xleftarrow{\hspace*{0.13\linewidth}}\text{CLIP guidance+DDNM} \xrightarrow{\hspace*{0.13\linewidth}}$\\
\includegraphics[width=0.151\linewidth]{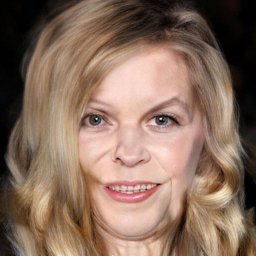}
\includegraphics[width=0.151\linewidth]{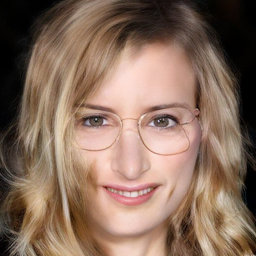}
\includegraphics[width=0.151\linewidth]{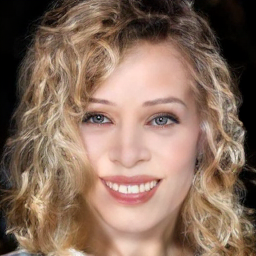}
\includegraphics[width=0.151\linewidth]{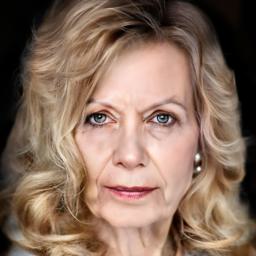}
\includegraphics[width=0.151\linewidth]{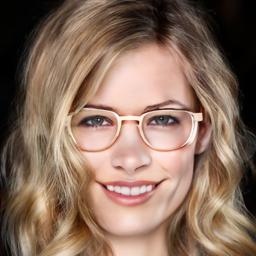}
\includegraphics[width=0.151\linewidth]{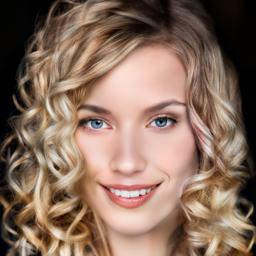}
 \\$\xleftarrow{\hspace*{0.17\linewidth}}\text{\cite{ramesh2022hierarchical}+DDNM} \xrightarrow{\hspace*{0.17\linewidth}}\xleftarrow{\hspace*{0.17\linewidth}}\text{\cite{saharia2022photorealistic}+DDNM} \xrightarrow{\hspace*{0.17\linewidth}}$\\
Text prompts: `Elderly woman', ` Smiling woman with glasses', `Woman with curly hair'\\

  \includegraphics[width=0.151\linewidth]{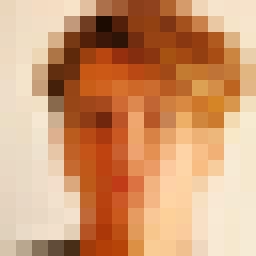}
\includegraphics[width=0.151\linewidth]{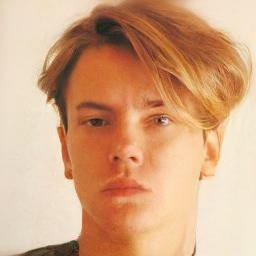}
\includegraphics[width=0.151\linewidth]{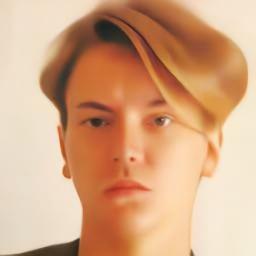}
\includegraphics[width=0.151\linewidth]{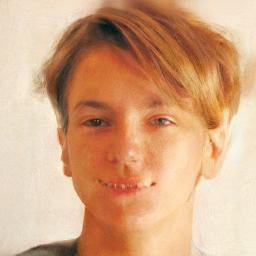}
\includegraphics[width=0.151\linewidth]{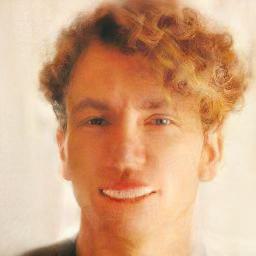}
\includegraphics[width=0.151\linewidth]{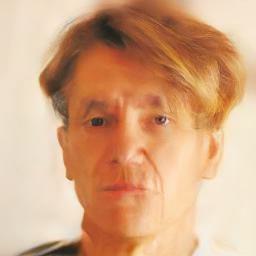}\\
\phantom{AAA}\footnotesize{LR}\phantom{AAAAAAAAAA}\footnotesize{DPS}\cite{chung2022diffusion}\phantom{AAAAAAAAAAA}DDNM\cite{wang2022zero}\phantom{AAA}$\xleftarrow{\hspace*{0.13\linewidth}}\text{CLIP guidance+DDNM} \xrightarrow{\hspace*{0.13\linewidth}}$\\
\includegraphics[width=0.151\linewidth]{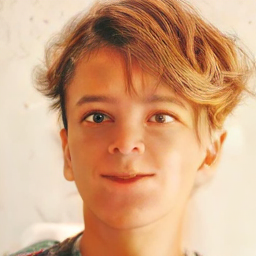}
\includegraphics[width=0.151\linewidth]{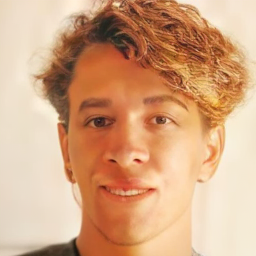}
\includegraphics[width=0.151\linewidth]{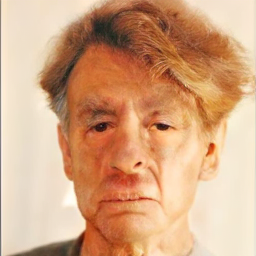}
\includegraphics[width=0.151\linewidth]{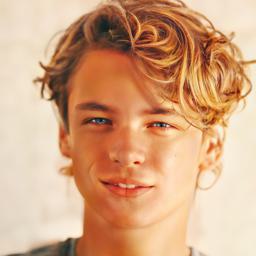}
\includegraphics[width=0.151\linewidth]{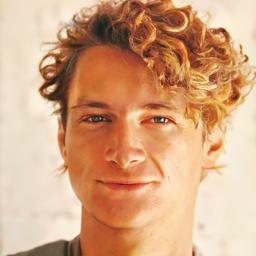}
\includegraphics[width=0.151\linewidth]{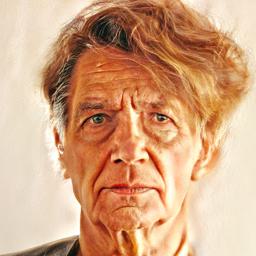}
 \\$\xleftarrow{\hspace*{0.17\linewidth}}\text{\cite{ramesh2022hierarchical}+DDNM} \xrightarrow{\hspace*{0.17\linewidth}}\xleftarrow{\hspace*{0.17\linewidth}}\text{\cite{saharia2022photorealistic}+DDNM} \xrightarrow{\hspace*{0.17\linewidth}}$\\
Text prompts: `Boy', ` Smiling man with curly hair', `Elderly man'\\
\caption{Exploring solutions for $16\times$ SR of face images.\label{fig:supple16xfaces_2}}
\end{figure*}
    \begin{figure*}[htb]
\centering
\scriptsize
\hspace{0.05\linewidth}{\bf LR input}\hspace{0.03\linewidth}
$\xleftarrow{\hspace*{0.14\linewidth}}\text{\bf Imagen$-$DDNM} \xrightarrow{\hspace*{0.14\linewidth}}\xleftarrow{\hspace*{0.14\linewidth}}\text{\bf unCLIP$-$DDNM} \xrightarrow{\hspace*{0.14\linewidth}}$\\
  \hspace{-2pt}\includegraphics[width=0.14\linewidth]{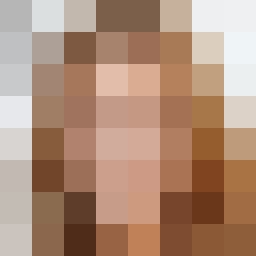}
\hspace{-2pt}\includegraphics[width=0.14\linewidth]{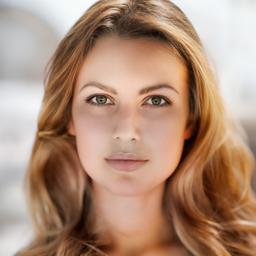}
\hspace{-2pt}\includegraphics[width=0.14\linewidth]{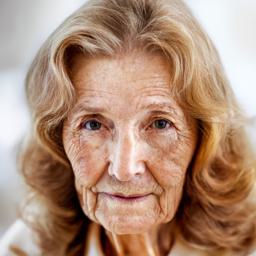}
\hspace{-2pt}\includegraphics[width=0.14\linewidth]{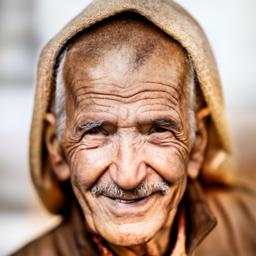}
\hspace{-2pt}\includegraphics[width=0.14\linewidth]{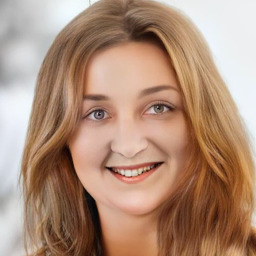}
\hspace{-2pt}\includegraphics[width=0.14\linewidth]{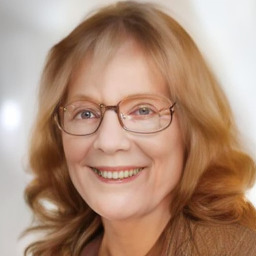}
\hspace{-2pt}\includegraphics[width=0.14\linewidth]{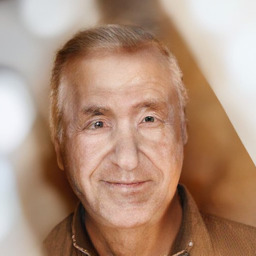}\\
Text prompts: `woman', `elderly woman', `elderly smiling man'\\
  \hspace{-2pt}\includegraphics[width=0.14\linewidth]{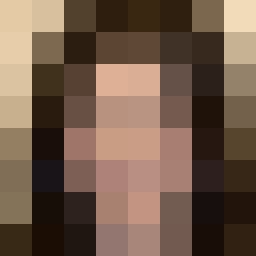}
\hspace{-2pt}\includegraphics[width=0.14\linewidth]{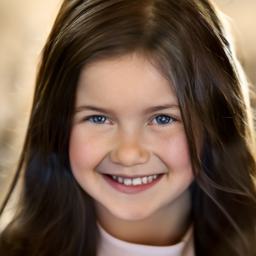}
\hspace{-2pt}\includegraphics[width=0.14\linewidth]{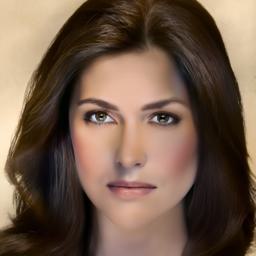}
\hspace{-2pt}\includegraphics[width=0.14\linewidth]{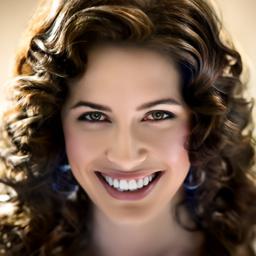}
\hspace{-2pt}\includegraphics[width=0.14\linewidth]{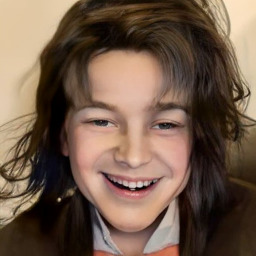}
\hspace{-2pt}\includegraphics[width=0.14\linewidth]{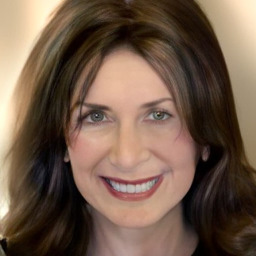}
\hspace{-2pt}\includegraphics[width=0.14\linewidth]{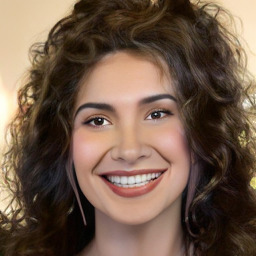}\\
Text prompts: `smiling child', `woman', `smiling woman with curly hair'\\
  \hspace{-2pt}\includegraphics[width=0.14\linewidth]{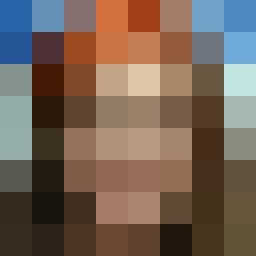}
\hspace{-2pt}\includegraphics[width=0.14\linewidth]{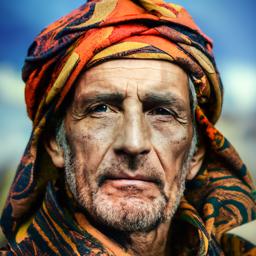}
\hspace{-2pt}\includegraphics[width=0.14\linewidth]{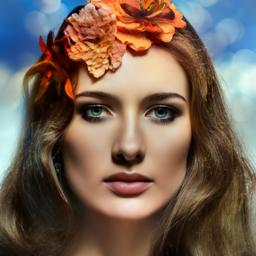}
\hspace{-2pt}\includegraphics[width=0.14\linewidth]{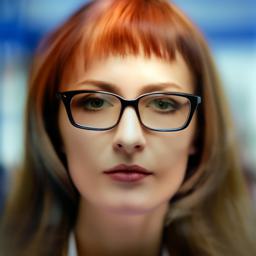}
\hspace{-2pt}\includegraphics[width=0.14\linewidth]{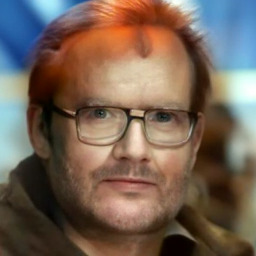}
\hspace{-2pt}\includegraphics[width=0.14\linewidth]{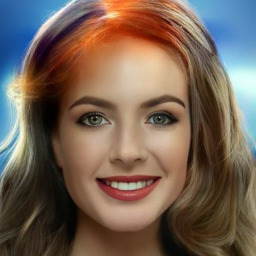}
\hspace{-2pt}\includegraphics[width=0.14\linewidth]{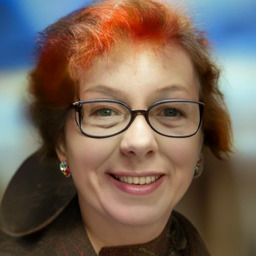}\\
Text prompts: `man', `young woman', `woman with glasses'\\ 
  \hspace{-2pt}\includegraphics[width=0.14\linewidth]{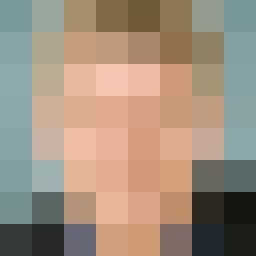}
\hspace{-2pt}\includegraphics[width=0.14\linewidth]{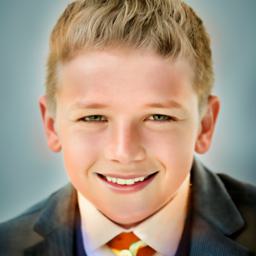}
\hspace{-2pt}\includegraphics[width=0.14\linewidth]{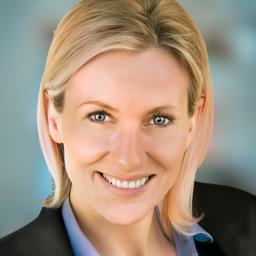}
\hspace{-2pt}\includegraphics[width=0.14\linewidth]{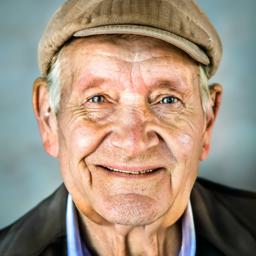}
\hspace{-2pt}\includegraphics[width=0.14\linewidth]{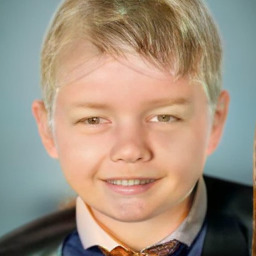}
\hspace{-2pt}\includegraphics[width=0.14\linewidth]{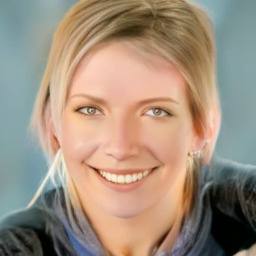}
\hspace{-2pt}\includegraphics[width=0.14\linewidth]{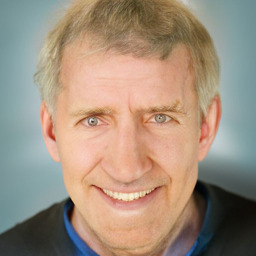}\\
Text prompts: `smiling boy', `smiling woman', `elderly smiling man'\\
  \hspace{-2pt}\includegraphics[width=0.14\linewidth]{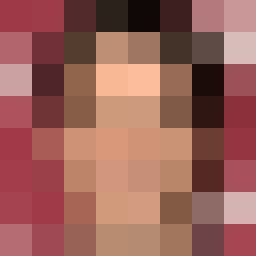}
\hspace{-2pt}\includegraphics[width=0.14\linewidth]{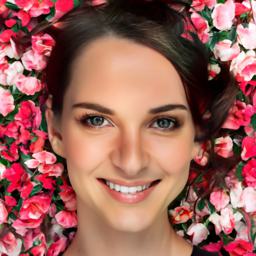}
\hspace{-2pt}\includegraphics[width=0.14\linewidth]{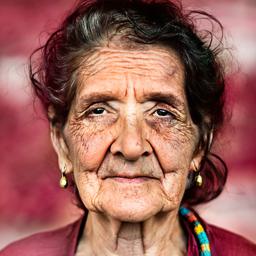}
\hspace{-2pt}\includegraphics[width=0.14\linewidth]{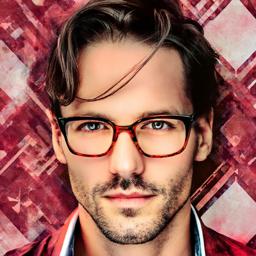}
\hspace{-2pt}\includegraphics[width=0.14\linewidth]{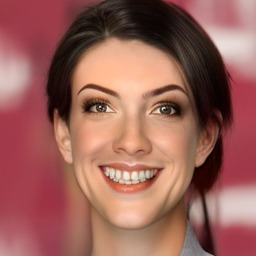}
\hspace{-2pt}\includegraphics[width=0.14\linewidth]{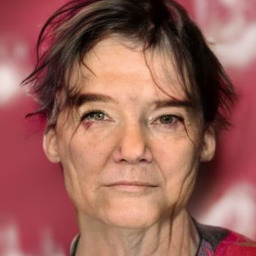}
\hspace{-2pt}\includegraphics[width=0.14\linewidth]{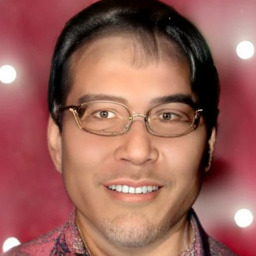}\\
Text prompts: `smiling woman',  `elderly woman', `man with glasses'\\
  \hspace{-2pt}\includegraphics[width=0.14\linewidth]{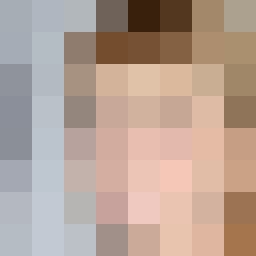}
  \hspace{-2pt}\includegraphics[width=0.14\linewidth]{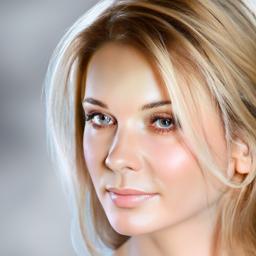}
\hspace{-2pt}\includegraphics[width=0.14\linewidth]{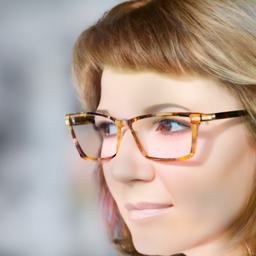}
\hspace{-2pt}\includegraphics[width=0.14\linewidth]{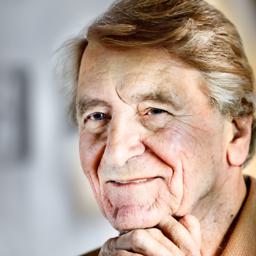}\hspace{-2pt}\includegraphics[width=0.14\linewidth]{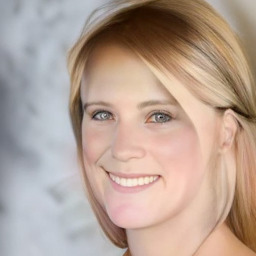}
\hspace{-2pt}\includegraphics[width=0.14\linewidth]{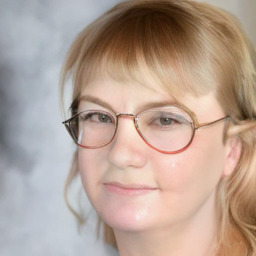}
\hspace{-2pt}\includegraphics[width=0.14\linewidth]{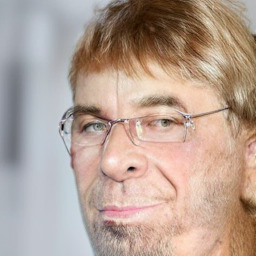}\\
Text prompts: `woman', `woman with glasses', `elderly smiling man'\\
  \hspace{-2pt}\includegraphics[width=0.14\linewidth]{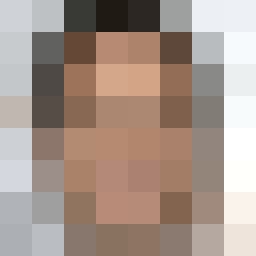}
\hspace{-2pt}\includegraphics[width=0.14\linewidth]{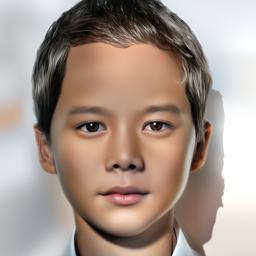}
\hspace{-2pt}\includegraphics[width=0.14\linewidth]{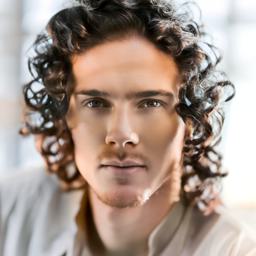}
\hspace{-2pt}\includegraphics[width=0.14\linewidth]{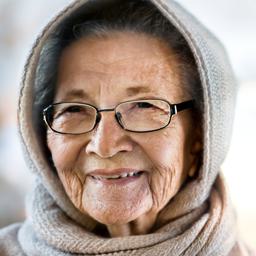}
\hspace{-2pt}\includegraphics[width=0.14\linewidth]{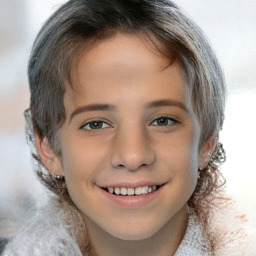}
\hspace{-2pt}\includegraphics[width=0.14\linewidth]{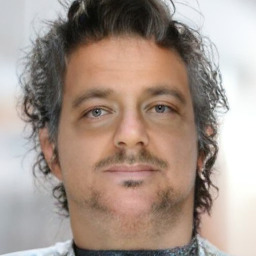}
\hspace{-2pt}\includegraphics[width=0.14\linewidth]{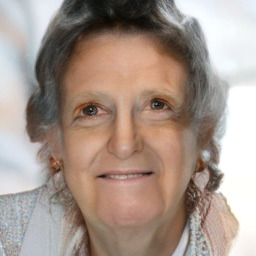}\\
Text prompts: `boy', `man with curly hair', `elderly smiling woman'\\
  \hspace{-2pt}\includegraphics[width=0.14\linewidth]{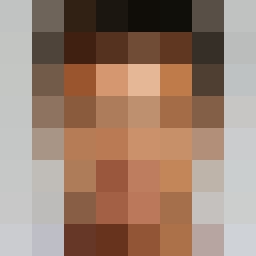}
\hspace{-2pt}\includegraphics[width=0.14\linewidth]{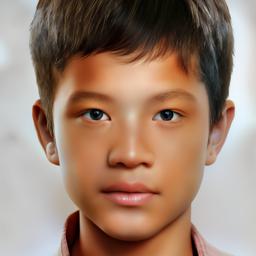}
\hspace{-2pt}\includegraphics[width=0.14\linewidth]{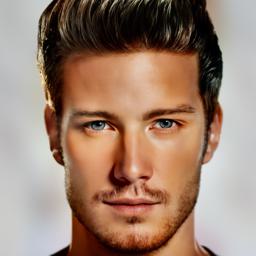}
\hspace{-2pt}\includegraphics[width=0.14\linewidth]{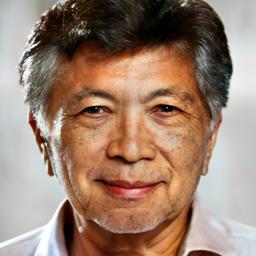}
\hspace{-2pt}\includegraphics[width=0.14\linewidth]{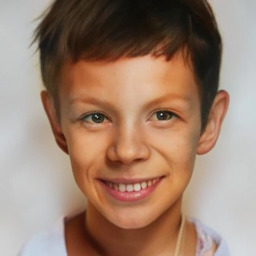}
\hspace{-2pt}\includegraphics[width=0.14\linewidth]{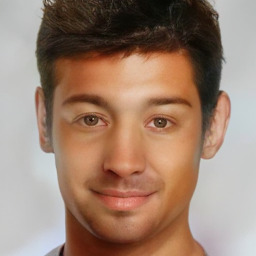}
\hspace{-2pt}\includegraphics[width=0.14\linewidth]{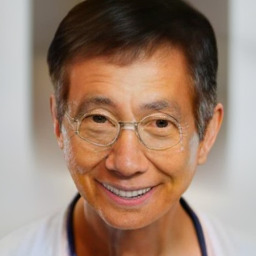}\\
Text prompts: `boy', `man', `elderly smiling man'
\caption{Exploring solutions for $32\times$ SR of face images.\label{fig:supple32xfaces}}
\end{figure*}
\section{Additional Results}\label{sec:res_supple}
\subsection{Open domain Image super-resolution}We provide additional qualitative results of open domain text guided super-resolution in \cref{fig:supple_nocaps_sr,fig:supple2_nocaps_sr}. While the solutions provided by DPS \cite{chung2022diffusion} are consistent with the measurements, the lack of additional information to guide the reconstruction results in unsatisfactory results while recovering complex scene content. The use of CLIP guidance with DDNM improves results over vanilla DPS, and \ti-DDNM improves adherence to text even further. Imagen-\cite{saharia2022photorealistic}+DDNM can nearly recover images adhering to complex prompts involving scene text, for instance, `A man with salt and pepper hair, a tie, and glasses is sitting behind a table with a sign that says Mexico in front of him.', or expression `A red panda is sitting on a tree squeezing its eyes shut and sticking out its tongue'.  It is also better at compositionality, for instance for text prompt `The woman is wearing a colorful kimono and carrying a floral print purple umbrella', Imagen-\cite{saharia2022photorealistic}+DDNM  can recover a floral print purple umbrella, where as unCLIP\cite{ramesh2022hierarchical}+DDNM recovers a purple umbrella without floral patterns, which appear instead on the woman's head and clothing.
\subsection{Exploring SR solutions through text} We provide qualitative results on exploring the solutions to face super-resolution to upsampling factors $8\times,~16\times$ and $32\times$.
\cref{fig:8xsr_supple,fig:8xsr_supple_2} show examples of exploring the solutions of $8\times$ super-resolution. The reconstructions of DPS~\cite{chung2022diffusion}, DDNM~\cite{wang2022zero} and \ti-DDNM methods are compared. The results demonstrate improved diversity using \ti-DDNM by exploring a variety of attributes. In contrast, when compared to vanilla DPS and DDNM which exhibit limited diversity. Similarly in \cref{fig:supple16xfaces,fig:supple16xfaces_2}  and \cref{fig:supple32xfaces} we illustrate examples of exploring the solutions of $16\times$ and $32\times$ super-resolution respectively. We can see in \cref{fig:supple32xfaces} that severe ill-posedness of $32\times$ SR task allows a wide range of solutions with varying personal attributes such as perceived age, gender, race, accessories etc.

\end{document}